%% file: root.tex
\documentclass[lettersize,journal]{IEEEtran}
\usepackage{amsmath,amsfonts}
\usepackage{amssymb}  % assumes amsmath package installed
\usepackage{algorithmic}
\usepackage{algorithm}
\usepackage{array}
\usepackage[caption=false,font=normalsize,labelfont=sf,textfont=sf]{subfig}
\usepackage{textcomp}
\usepackage{stfloats}
\usepackage{url}
\usepackage{verbatim}
\usepackage{graphicx}
\usepackage{graphics}
\usepackage{cite}
\usepackage{epsfig} % for postscript graphics files
\usepackage{times} % assumes new font selection scheme installed=
\usepackage{float}
\usepackage{booktabs}
\usepackage{caption}
\usepackage{lipsum}
\usepackage{multirow}
\usepackage{xcolor} 
\usepackage{subcaption}
\usepackage{amsmath}
\usepackage{amssymb}
\usepackage{xspace}
\usepackage{threeparttable}
\usepackage{enumitem}
\usepackage{dblfnote}
\usepackage{blindtext}
\usepackage{hyperref}
\usepackage{array}
\usepackage[table]{xcolor}
\usepackage{bbm}
\begin{document}
\include{lib}

\def\etal{et al.}
\newcommand{\rep}{DexRep\xspace}
\newcommand{\occ}{f_o}
\newcommand{\surf}{f_s}
\newcommand{\loc}{f_l}
\newcommand{\lqt}[1]{\textcolor{cyan}{#1}}
\newcommand{\lqtt}[1]{\textcolor{black}{#1}}
\newcommand{\lqttd}[1]{\textcolor{red}{TODO:#1}}
\newcommand{\polish}[1]{\textcolor{cyan}{#1}}

\newcommand\blfootnote[1]{% 
\begingroup 
\renewcommand\thefootnote{}\footnote{#1}% 
\addtocounter{footnote}{-1}% 
\endgroup 
}

\title{DexRepNet++: Learning Dexterous Robotic Manipulation with Geometric and Spatial Hand-Object Representations}

\author{Qingtao Liu, Zhengnan Sun, Yu Cui, Haoming Li, 
Gaofeng Li,~\IEEEmembership{Member,~IEEE,} Lin Shao, Jiming Chen,~\IEEEmembership{Fellow,~IEEE}, Qi Ye,~\IEEEmembership{Member,~IEEE}
        % <-this % stops a space
\thanks{This paper was produced by the IEEE Publication Technology Group. They are in Piscataway, NJ.}% <-this % stops a space
\thanks{Manuscript received April 19, 2021; revised August 16, 2021.}}

\markboth{Journal of \LaTeX\ Class Files,~Vol.~14, No.~8, August~2021}%
{Shell \MakeLowercase{\textit{et al.}}: A Sample Article Using IEEEtran.cls for IEEE Journals}



\twocolumn[{%
\renewcommand\twocolumn[1][]{#1}%
\maketitle
\vspace{-0.3cm}
\begin{center}
    % \centering
    \captionsetup{type=figure}
    \includegraphics[width=0.96\textwidth]{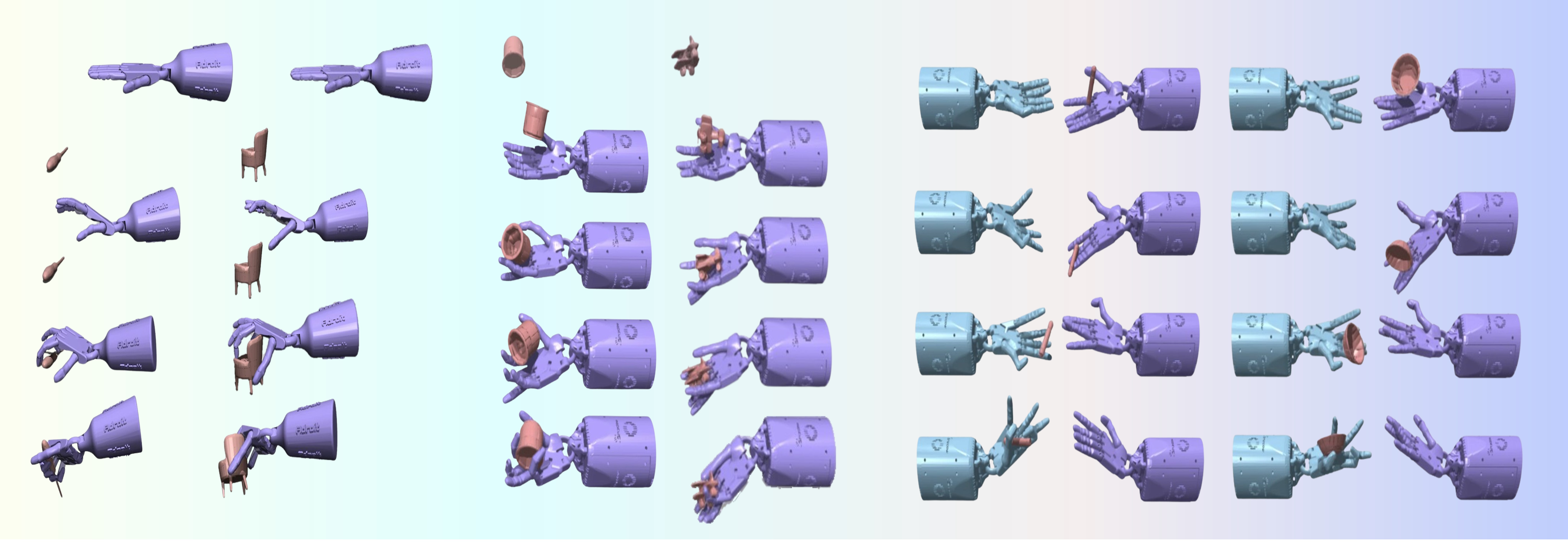}
    \captionof{figure}{Policies learned with our hand-object representation perform grasping, in-hand reorientation, and handover tasks with five-fingered dexterous robotic hands.}
    \label{fig:teaser}
\end{center}%
}]

\renewcommand{\thefootnote}{}
\footnotetext{This research is supported in part by the National Natural Science Foundation of China (Grant Number: 62088101, 62233013, 62293511), Key Research and Development Program of Zhejiang Province (No: 2025C01072). This research is also supported by the Ministry of Education, Singapore, under the Academic Research Fund Tier 1 (FY2024). (Corresponding author: Qi Ye.)}
\footnotetext{Qi Ye, Gaofeng Li, Jiming Chen are with the College of Control Science and Engineering and the State Key Laboratory of Industrial Control Technology, Zhejiang University, China (e-mail: \texttt{qi.ye@zju.edu.cn}, \texttt{gaofeng.li@zju.edu.cn}, \texttt{cjm@zju.edu.cn}).}
\footnotetext{Qingtao Liu, Zhengnan Sun, and Yu Cui are with the College of Control Science and Engineering, Zhejiang University, Hangzhou 310027, China (e-mail: \texttt{l\_qingtao@zju.edu.cn}, \texttt{zn.sun@zju.edu.cn}, \texttt{cuiyu0627@163.com})}

% \footnotetext{ is with the College of Control Science and Engineering and the State Key Laboratory of Industrial Control Technology, Zhejiang University, and also with the Key Key Lab of CS\&AUS of Zhejiang Province, China (e-mail: qi.ye@zju.edu.cn).}
\footnotetext{Lin Shao is with the Department of Computer Science, National University of Singapore, 15 Computing Dr, 117418, Singapore (e-mail: \texttt{linshao@nus.edu.sg}).}
% \footnotetext{ is with the State Key Laboratory of Industrial Control Technology,
% Department of Control, Zhejiang University, Hangzhou 310027, China (e-mail: cjm@zju.edu.cn).}
\footnotetext{Project page: \href{https://lqts.github.io/DexRepNet2/}{https://lqts.github.io/DexRepNet2/}}
\renewcommand{\thefootnote}{\arabic{footnote}}

%%%%%%%%%%%%%%%%%%%%%%%%%%%%%%%%%%%%%%%%%%%%%%%%%%%%%%%%%%%%%%%%%%%%%%%%%%%%%%%%

\input{txt/abstract.tex}
%%%%%%%%%%%%%%%%%%%%%%%%%%%%%%%%%%%%%%%%%%%%%%%%%%%%%%%%%%%%%%%%%%%%%%%%%%%%%%%%

\input{txt/introduction.tex}
\input{txt/related_work.tex}
\input{txt/method.tex}

\input{txt/exper_result.tex}

\input{txt/conclusion.tex}

\bibliographystyle{IEEEtran}
% \bibliography{IEEEabrv, IEEEexample,mybibfile}
\bibliography{bibfiles}

\vspace{-0.5cm}
\begin{IEEEbiography}
[{\includegraphics[width=1in,height=1.25in,clip,keepaspectratio]{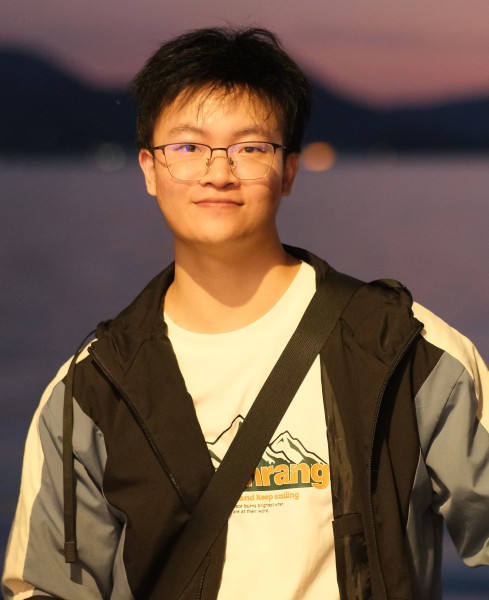}}]{Qingtao Liu} is pursuing a PhD at the College of Control Science and Engineering, Zhejiang University, under the supervision of Qi Ye and Jiming Chen. He graduated from China University of Geosciences (Wuhan) with a bachelor's degree in 2021. His current research mainly focuses on dexterous manipulation and multimodal representation learning.
\end{IEEEbiography}
\vspace{-1.0cm}
\begin{IEEEbiography}
[{\includegraphics[width=1in,height=1.25in,clip,keepaspectratio]{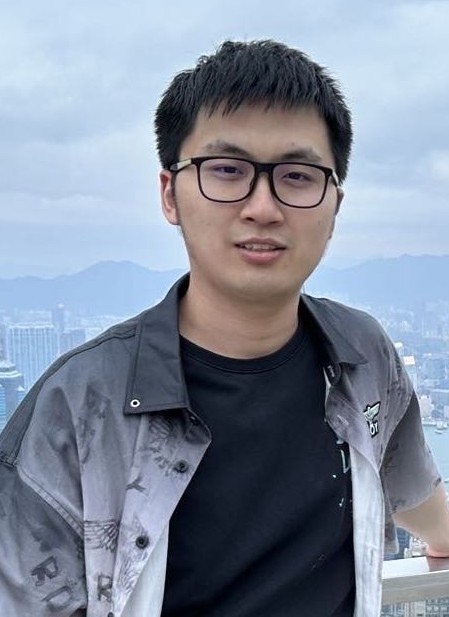}}]{Zhengnan Sun} received the B.S. degree from the College of Electronic Engineering of Zhejiang University, Hangzhou, China, in 2023. He is currently pursuing the Master's degree with the College of Control Science and Engineering of Zhejiang University, Hangzhou, China. His research interests lie in robotic dexterous hands and multi-modal fusion.
\end{IEEEbiography}

\vspace{-1cm}
\begin{IEEEbiography}
[{\includegraphics[width=1in,height=1.25in,clip,keepaspectratio]{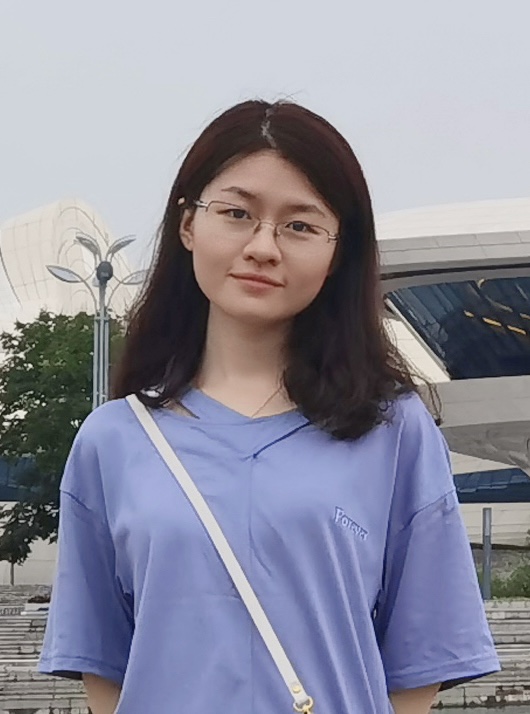}}]{Yu Cui} is currently pursuing a Master's degree at the College of Control Science and Engineering, Zhejiang University. Before that, she obtained her Bachelor's degree from University of Science and Technology Beijing. Her research interests focus on robot learning and dexterous manipulation.
\end{IEEEbiography}
\vspace{-1cm}
\begin{IEEEbiography}
[{\includegraphics[width=1in,height=1.25in,clip,keepaspectratio]{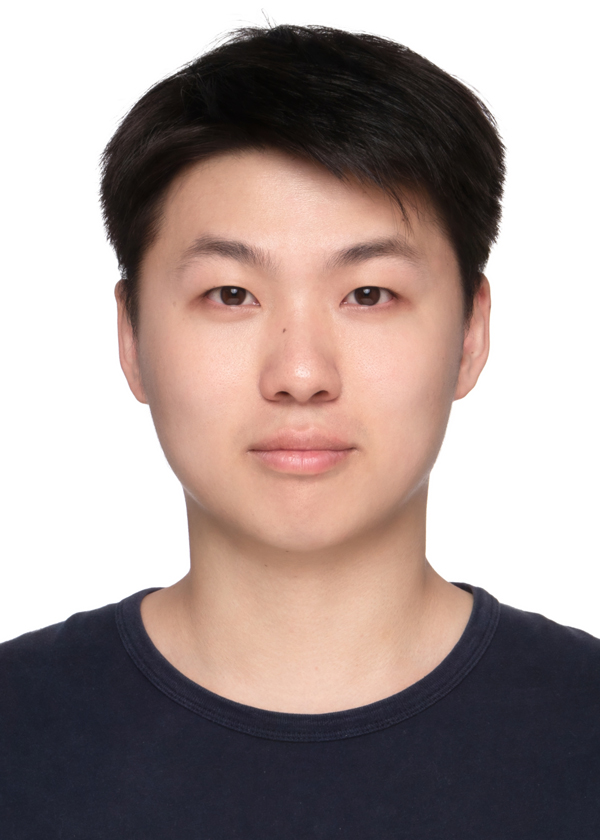}}]{Haoming Li} is currently pursuing a Ph.D. degree at the College of Control Science and Engineering, Zhejiang University. Before this, he obtained both his Bachelor's and Master's degrees from Shenzhen University. His research interests primarily focus on dexterous hand manipulation skill generation and motion planning.
\end{IEEEbiography}
\vspace{-1cm}

\begin{IEEEbiography}[{\includegraphics[width=1in,height=1.25in,clip,keepaspectratio]{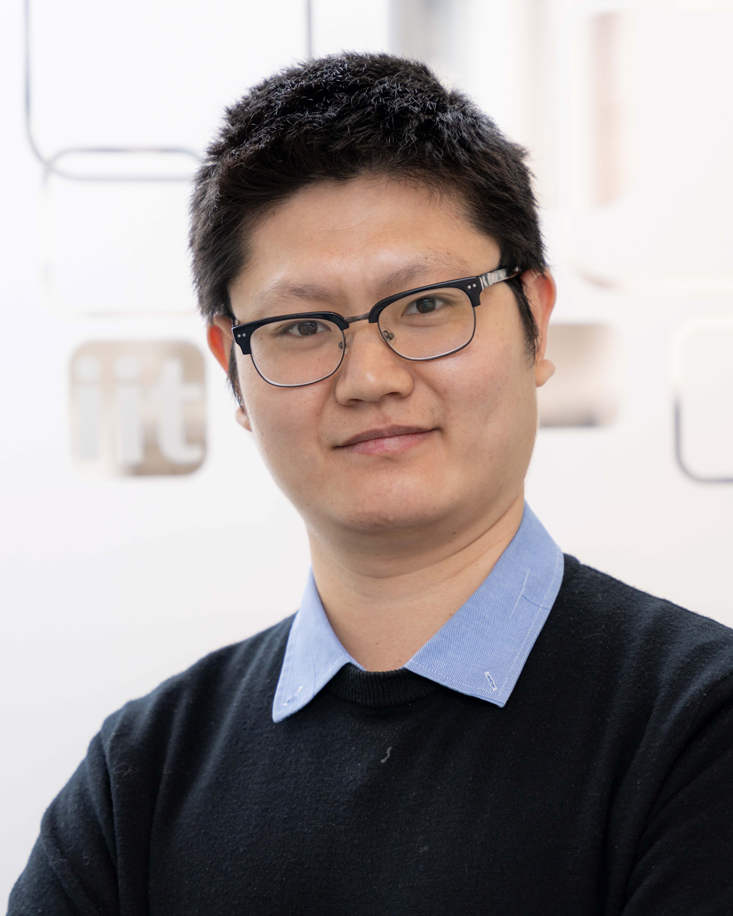}}]{Gaofeng Li} (Member, IEEE)
is currently a tenure-tracked professor under the Hundred Talents Program at Zhejiang University. He is the founder and director of the ARTs (Agile Robotic Tele-systems) Lab. His research interests include Lie Group in Robotics, Robotic Manipulation, Haptic Teleoperation, Imitation Learning, and Soft Robotics.
He serves/served as a lead guest editor for the International Journal of Humanoid Robotics (IJHR), and the Late Breaking Report Chair for the Organizing Committee of IEEE RO-MAN 2024. He is also an independent Reviewer for IJRR, IEEE T-RO, IEEE T-ASE, IEEE/ASME T-Mech, IEEE RAM, IEEE RA-L, IEEE ICRA, IEEE IROS, etc.
\end{IEEEbiography}
\vspace{-0.5cm}
\begin{IEEEbiography}[{\includegraphics[width=1in,height=1.25in,clip,keepaspectratio]{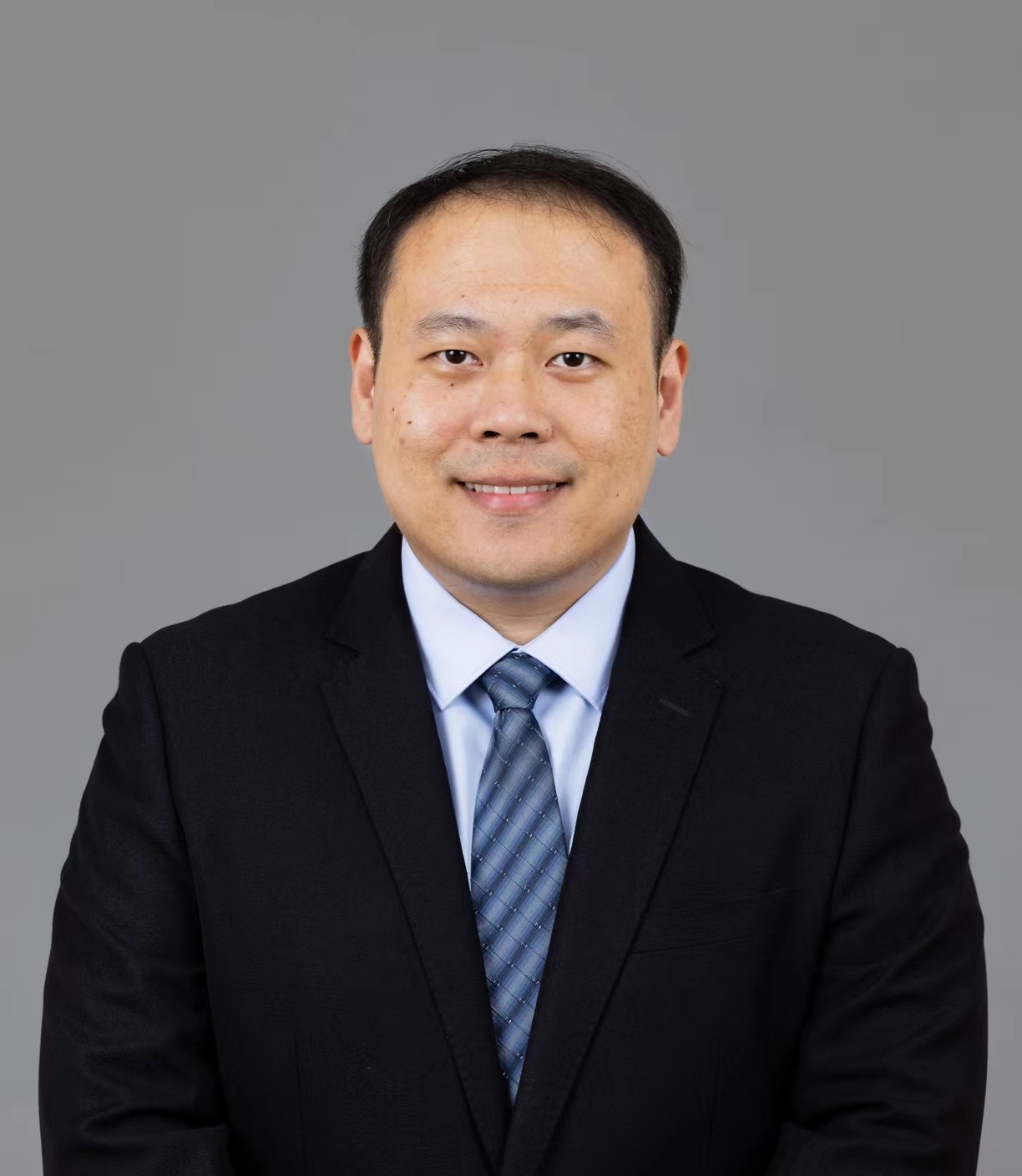}}]{Lin Shao}
 is an Assistant Professor in the Department of Computer Science at the National University of Singapore (NUS), School of Computing. His research interests lie at the intersection of Robotics and Artificial Intelligence. His long-term goal is to build general-purpose robotic systems that intelligently perform a diverse range of tasks in a large variety of environments in the physical world. Specifically, his group is interested in developing algorithms and systems to provide robots with the abilities of perception and manipulation.   He is a co-chair of the Technical Committee on Robot Learning in the IEEE Robotics and Automation Society and serves as the Associated Editor at ICRA 2024. His work received the Best System Paper Award finalist at RSS 2023. Previously, he received his PhD at Stanford University, advised by Jeannette Bohg. He received his BS from Nanjing University.
\end{IEEEbiography}
\vspace{-0.5cm}
\begin{IEEEbiography}
[{\includegraphics[width=1in,height=1.25in,clip,keepaspectratio]{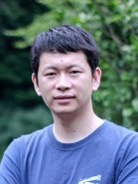}}]{Jiming Chen} is a Changjiang Scholars Professor with College of control science and engineering, Zhejiang University.  Currently, he serves/served associate editors for ACM TECS, IEEE TPDS, IEEE Network, IEEE TCNS, IEEE TII, etc. He has been appointed as a distinguished lecturer of IEEE Vehicular Technology Society 2015, and selected in National Program for Special Support of Top-Notch Young Professionals, and also funded Excellent Youth Foundation of NSFC. He also was the recipients of IEEE INFOCOME 2014 Best Demo Award, IEEE ICCC 2014 best paper award, IEEE PIMRC 2012 best paper award, and JSPS Visiting Fellowship 2011. He also received the IEEE Comsoc Asia-pacific Outstanding Young Researcher Award 2011. He is a Distinguished Lecturer of IEEE Vehicular Technology Society (2015-2018), and a Fellow of IEEE. His research interests include networked control, sensor networks, cyber security, IoT.
\end{IEEEbiography}
\vspace{-1cm}
\begin{IEEEbiography}
[{\includegraphics[width=1in,height=1.25in,clip,keepaspectratio]{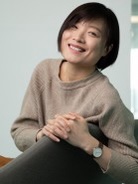}}]{Qi Ye} is a Tenure-Track Professor under the Hundred Talents Program at Zhejiang University. Before joining Zhejiang University, she was a research scientist of Mixed Reality \& AI Lab at Cambridge, Microsoft and obtained Ph.D. degree at Imperial College London. She is interested in and working on human-computer interaction, particularly vision understanding involving hands, and a more general setting where humans interact with the environment. She is passionate about 3D vision and its applications, particularly in Mixed/Augmented/Virtual Reality and automatic control.
\end{IEEEbiography}


\vfill

\end{document}

%% file: lib.tex
\newcommand{\yq}[1]{\textcolor{red}{\textbf{YQ: #1}}}
\newcommand{\reflabel}{dummy} % Dummy initial reflabel - use renewcommand...

% \newcommand{\be}{\begin{equation}}
% \newcommand{\ee}{\end{equation}}
% \newcommand{\eqlabel}[1]{\label{eq:\reflabel-#1}}
% \renewcommand{\eqref}[2][\reflabel]{(\ref{eq:#1-#2})}

% Generic reference commands
\newcommand{\seclabel}[1]{\label{sec:\reflabel-#1}}
\newcommand{\secref}[2][\reflabel]{Section~\ref{sec:#1-#2}}
\newcommand{\Secref}[2][\reflabel]{Section~\ref{sec:#1-#2}}
\newcommand{\secrefs}[3][\reflabel]{Sections~\ref{sec:#1-#2} and~\ref{sec:#1-#3}}

\newcommand{\eqlabel}[1]{\label{eq:\reflabel-#1}}
\renewcommand{\eqref}[2][\reflabel]{(\ref{eq:#1-#2})}
\newcommand{\Eqref}[2][\reflabel]{(\ref{eq:#1-#2})}
\newcommand{\eqrefs}[3][\reflabel]{(\ref{eq:#1-#2}) and~(\ref{eq:#1-#3})}

\newcommand{\figlabel}[2][\reflabel]{\label{fig:#1-#2}}
\newcommand{\figref}[2][\reflabel]{Fig.~\ref{fig:#1-#2}}
\newcommand{\Figref}[2][\reflabel]{Fig.~\ref{fig:#1-#2}}
\newcommand{\figsref}[3][\reflabel]{Figs.~\ref{fig:#1-#2} and~\ref{fig:#1-#3}}
\newcommand{\Figsref}[3][\reflabel]{Figs.~\ref{fig:#1-#2} and~\ref{fig:#1-#3}}

\newcommand{\tablelabel}[2][\reflabel]{\label{table:#1-#2}}
\newcommand{\tableref}[2][\reflabel]{Table~\ref{table:#1-#2}}
\newcommand{\Tableref}[2][\reflabel]{Table~\ref{table:#1-#2}}
\newcommand{\etal}{et al.}
\newcommand{\eg}{e.g. }
\newcommand{\ie}{i.e. }
\newcommand{\etc}{etc. }

%%%
%%% Stuff for bold maths typesetting  ----------------------------------------
%%%
%%%   e.g. use "\bfmu" for boldface mu symbol
%%%
\def\bfmu{\mbox{\boldmath$\mu$}}
\def\bftau{\mbox{\boldmath$\tau$}}
\def\bftheta{\mbox{\boldmath$\theta$}}
\def\bfdelta{\mbox{\boldmath$\delta$}}
\def\bfphi{\mbox{\boldmath$\phi$}}
\def\bfpsi{\mbox{\boldmath$\psi$}}
\def\bfeta{\mbox{\boldmath$\eta$}}
\def\bfnabla{\mbox{\boldmath$\nabla$}}
\def\bfGamma{\mbox{\boldmath$\Gamma$}}

%%% Make figure placement a little more predictable.
% We trust the user to move figures if this results
% in ugliness.
% Minimize bad page breaks at figures
%\renewcommand{\textfraction}{0.01}
%\renewcommand{\floatpagefraction}{0.99}
%\renewcommand{\topfraction}{0.99}
%\renewcommand{\bottomfraction}{0.99}
%\renewcommand{\dblfloatpagefraction}{0.99}
%\renewcommand{\dbltopfraction}{0.99}
%\setcounter{totalnumber}{99}
%\setcounter{topnumber}{99}
%\setcounter{bottomnumber}{99}
%
%% Add a period to the end of an abbreviation unless there's one
%% already, then \xspace.
%\makeatletter
%\DeclareRobustCommand\onedot{\futurelet\@let@token\@onedot}
%\def\@onedot{\ifx\@let@token.\else.\null\fi\xspace}
%
%\def\eg{\emph{e.g}\onedot} \def\Eg{\emph{E.g}\onedot}
%\def\ie{\emph{i.e}\onedot} \def\Ie{\emph{I.e}\onedot}
%\def\cf{\emph{c.f}\onedot} \def\Cf{\emph{C.f}\onedot}
%\def\etc{\emph{etc}\onedot} \def\vs{\emph{vs}\onedot}
%\def\wrt{w.r.t\onedot} \def\dof{d.o.f\onedot}
%\def\etal{\emph{et al}\onedot}
%\makeatother

% ---------------------------------------------------------------

\newcommand{\R}{\mathbb{R}}

\newcommand{\be}{\begin{equation}}
\newcommand{\ee}{\end{equation}}

%% file: txt/abstract.tex
\begin{abstract}

Robotic dexterous manipulation is a challenging problem due to high degrees of freedom (DoFs) and complex contacts of multi-fingered robotic hands. Many existing deep reinforcement learning (DRL) based methods aim at improving sample efficiency in high-dimensional output action spaces.
However, existing works often overlook the role of representations in achieving generalization
% less attention has been paid to the representations for the generalization 
of a manipulation policy in the complex input space during the hand-object interaction.  
In this paper, we propose \rep, a novel hand-object interaction representation to capture object surface features and spatial relations between hands and objects for dexterous manipulation skill learning.  Based on \rep, policies are learned for three dexterous manipulation tasks, \ie grasping, in-hand reorientation, bimanual handover, and extensive experiments are conducted to verify the effectiveness. In simulation, for grasping, the policy learned with 40 objects achieves a success rate of 87.9\% on more than 5000 unseen objects of diverse categories, significantly surpassing existing work trained with thousands of objects; for the in-hand reorientation and handover tasks, the policies also boost the success rates and other metrics of existing hand-object representations by 20\% to 40\%. The grasp policies with \rep are deployed to the real world under multi-camera and single-camera setups and demonstrate a small sim-to-real gap.

\end{abstract}


%% file: txt/introduction.tex
\section{Introduction}

\IEEEPARstart{D}{exterous} manipulation is a fundamental capability for enabling robots to perform complex, human-level tasks in unstructured environments. Compared to parallel grippers, anthropomorphic multi-fingered hands provide superior flexibility and control, making them particularly well-suited for handling diverse objects.

Recent progress~\cite{DAPG:rajeswaran2017learning,ILAD:wu2022learning,qin2022dexmv,bao2023dexart,mandikal2022dexvip,xu2023unidexgrasp,wang2023DexGraspNet,wan2023unidexgrasp++,taochen2023visualdexterity,bi-dexhands,li2023dexdeform} has highlighted the potential of reinforcement learning (RL) for acquiring dexterous manipulation skills with multi-fingered hands. However, two major challenges continue to limit the generalization and scalability of existing methods: (i)~the high dimensionality of the control space, and (ii)~the complex, highly variable nature of hand-object interactions. Dexterous hands typically have 16-24 degrees of freedom (DoFs)~\cite{allegrohand,shaw2023leap,shadowrobot_hand}, resulting in a large continuous action space that makes policy learning inefficient and unstable. In parallel, the interaction dynamics between articulated fingers and objects with diverse geometries lead to variable and sensitive contact patterns, where even slight changes in object shape or orientation can require dramatically different hand configurations.

To address these difficulties, prior efforts have employed imitation learning~\cite{DAPG:rajeswaran2017learning, ILAD:wu2022learning, mandikal2022dexvip} and curriculum-based strategies~\cite{xu2023unidexgrasp, wan2023unidexgrasp++} to improve sample efficiency and training robustness. While these methods have achieved notable performance gains by scaling demonstrations or shaping learning progressions, they often overlook a more fundamental question: how hand-object interaction should be encoded to facilitate generalizable skill learning. Common representations include encoding object geometry via pretrained point-based networks (e.g., PointNet~\cite{qi2017pointnet}) and describing hand-object relations using absolute positions, joint angles, or relative distances. However, such representations are often tightly coupled to specific object instances and hand configurations, leading to poor generalization across novel objects or tasks.

Meanwhile, insights from grasp taxonomies~\cite{feix2016taxonomy} and human motor control~\cite{mason2001hand} suggest that hand-object interactions, despite their apparent complexity, often lie on low-dimensional manifolds. For instance, grasps can be clustered into a finite set of discrete types, and finger motions often follow coordinated synergies. These observations point toward the possibility of achieving better generalization by leveraging more structured and compact representations.

Motivated by these insights, we argue that the key to enabling generalizable dexterous manipulation lies not in increasing the scale of demonstrations or task engineering, but in rethinking the representation of hand-object interaction itself. Specifically, we propose to move beyond global shape encodings and instead develop a representation grounded in \emph{local surface geometry} and \emph{spatial proximity} between the hand and object. The underlying intuition is that many objects share common local structures (e.g., handles, lips, edges), and similar manipulation strategies can often be applied based on local geometry, regardless of the object's global shape. Coarse global features may suffice for approach and pre-shaping of grasp, whereas local geometric cues are essential for contact-rich actions such as grasp closure or in-hand adjustment. This motivates a hybrid representation that encodes both global structure and local detail in a unified framework.

Based on this motivation, we introduce \textbf{\rep}, a structured representation of hand-object interaction that encodes both spatial and geometric cues relative to the robot hand. \rep comprises three components:
\textbf{(1) Occupancy Feature:} a voxelized representation of the object surface captured from the hand's local frame;
\textbf{(2) Surface Feature:} a set of distances and surface normals between hand keypoints and their nearest points on the object;
\textbf{(3) Local-Geo Feature:} fine-grained geometric descriptors extracted from surface regions near potential contact areas using a pretrained PointNet~\cite{qi2017pointnet} to extend the surface normals in Surface Feature for more abundant local information.
The Occupancy feature captures global shape information seen by the approaching hands via a coarse occupancy volume instead of static point clouds carrying detailed geometry and the Surface Feature and Local-Geo Feature capture fine-grained local geometry: the combination of the coarse global and finer local features fully captures object surface information and also ensures generalization to unseen objects that share partial geometric similarities with the training set. Surface Feature and Local-Geo Feature capture the dynamics of the interaction and the geometry feature of the most related object surface area for each hand part in potential contacts. 
% Together, these features provide a compact yet expressive encoding of the interaction state, balancing coarse and fine information in a way that supports generalization to new objects, tasks, and hand morphologies.

We evaluate \rep on three manipulation tasks of increasing complexity: (1) object grasping, (2) in-hand reorientation, and (3) bimanual handover. These tasks cover both single- and dual-hand settings, and involve different requirements in terms of spatial precision, contact adaptation, and coordination. Furthermore, we show that \rep remains effective when computed from partial observations (e.g., single-view depth input), making it practical for real-world deployment without requiring full object observations.

Our empirical results demonstrate the following key advantages:
\textbf{(1) Strong Generalization:} A grasping policy trained on only 40 objects generalizes to over 5,000 unseen shapes with an average success rate of nearly 88\%.
\textbf{(2) Multi-task Adaptability:} Despite being task-agnostic, \rep achieves superior performance across multiple manipulation tasks—including grasping, in-hand reorientation, and handover—outperforming task-specific baselines.
\textbf{(3) Morphological Transferability:} The same representation supports different robotic hands, including 2-, 3-, 4-, and 5-finger configurations.
\textbf{(4) Robust Real-world Deployment:} Our system achieves up to 85\% grasping success using multi-view input, and over 75\% using only partial observations.
\textbf{(5) Small Sim-to-real Gap:} Simulation-trained policies transfer to hardware with less than 5\% drop in performance.

This paper extends our prior conference work~\cite{liu2023dexrepnet} in several key directions:
\begin{itemize}
    \item \textbf{Expanded Task Coverage:} We introduce new evaluations on in-hand manipulation and dual-arm handover, showcasing the versatility of \rep.
    \item \textbf{Robustness to Partial Observations:} We demonstrate successful deployment with only single-view depth input, increasing practical applicability.
    \item \textbf{Comprehensive Analysis:} We conduct detailed ablations on feature types, voxel resolution, and training object diversity, offering insights into design trade-offs.
    \item \textbf{Improved Baseline Comparison:} We scale up evaluations with stronger baselines, including UniDexGrasp++~\cite{wan2023unidexgrasp++}, and larger object sets.
    \item \textbf{Real-world Implementation:} We build a full real-world system combining Allegro Hand, Unitree Z1 arm, and Azure Kinect, and validate performance under practical conditions.
\end{itemize}

In summary, we propose \textbf{\rep}, a compact, structured representation of hand-object interaction that supports robust and generalizable learning for dexterous manipulation. \rep integrates global and local geometry, adapts to different hands and observations, and facilitates sim-to-real deployment. We believe this representation offers a promising foundation for advancing general-purpose manipulation in complex, real-world environments.

%% file: txt/related_work.tex
\section{Related Work}

\subsection{Dexterous Grasping and Manipulation}

Dexterous grasping and manipulation are core tasks for multi-fingered robotic hands. Grasping aims to establish a stable initial contact with the object, while manipulation extends this capability to dynamic control for object repositioning, reorientation, or handover. Despite their differences in execution, both tasks face shared challenges, including high-dimensional control, complex contact dynamics, and the need for generalization across diverse object geometries.

Early research approached these problems from a planning perspective, relying on precise models of hand kinematics and object dynamics~\cite{han1998dextrous, rus1999hand, mordatch2012contact, kumar2014real}. However, such methods are typically constrained by their reliance on accurate geometric and physical modeling, which limits their applicability in unstructured or dynamic environments.

Recent advances have shifted toward learning-based approaches, especially reinforcement learning (RL) and imitation learning (IL), to bypass explicit modeling. In grasping, methods like DAPG~\cite{DAPG:rajeswaran2017learning} and ILAD~\cite{ILAD:wu2022learning} leverage expert demonstrations to improve sample efficiency and stabilize training. GRAFF~\cite{GRAFF:mandikal2021learning} and DexVIP~\cite{mandikal2022dexvip} introduce affordance priors from human demonstrations to guide learning, while Christen \etal~\cite{Christen_2022_CVPR} generate grasp trajectories from static visual inputs. UniDexGrasp~\cite{xu2023unidexgrasp} and UniDexGrasp++~\cite{wan2023unidexgrasp++} further extend these ideas by incorporating diverse grasp configurations and curriculum learning strategies to enhance generalization.

For more complex manipulation tasks, recent works explore a wide range of skills—from tool use with chopsticks~\cite{yang2022chopsticks} to catching objects~\cite{lan2023dexcatch} and in-hand reorientation~\cite{huang2021generalization, chen2022inhand, qi2023general}. Large-scale vision-based approaches such as DexArt~\cite{bao2023dexart} and DexMV~\cite{qin2022dexmv} aim to boost generalization via diverse datasets and visual pretraining. Other efforts improve RL algorithms~\cite{omer2021model, he2022discovering} or adopt cross-modal imitation strategies from human videos~\cite{qin2022dexmv, qin2022one}.

Despite these advances, most existing methods focus on improving policy learning frameworks or scaling training data, while paying relatively less attention to how hand-object interaction is represented. Common strategies encode global object geometry~\cite{qi2017pointnet} or use position-based features (e.g., relative distances or joint signals). However, such representations often lack task-specific structure and fail to capture fine-grained contact variations across diverse objects. While a few recent efforts—e.g.,~\cite{huang2021generalization, she2022ibs}—explore structured representations, the focus remains limited.

In contrast, our work centers on developing a structured and task-relevant representation, \textbf{DexRep}, that explicitly encodes local surface geometry and hand-object spatial proximity. This design enables consistent application across multiple dexterous tasks—including grasping, in-hand manipulation, and handover—and supports generalization across novel objects and different robotic hands.

\subsection{Representation for Robotic Manipulation}

In RL-based manipulation, effective policy learning critically depends on the quality of state representation—particularly how the interaction between the hand and the environment is perceived and encoded. To this end, vision is widely used to represent object geometry and spatial context, leveraging various modalities such as RGB images~\cite{levine2018handeye}, depth maps~\cite{liu2020deep}, point clouds~\cite{ILAD:wu2022learning, wei2022dvgg, qin2023dexpoint}, meshes~\cite{varley2017shape}, or multimodal combinations~\cite{GRAFF:mandikal2021learning, mandikal2022dexvip, cao2021suctionnet}. More recently, visual pretraining has gained attention as a way to build general-purpose encoders for downstream tasks. R3M~\cite{nair2022r3m}, for example, learns from large-scale human video data, while RealMAE~\cite{radosavovic2023real} extends masked autoencoding~\cite{he2022mae} to real-world manipulation settings.

For encoding the hand, commonly used representations include joint angles~\cite{ILAD:wu2022learning, lan2023dexcatch, huang2021generalization}, hand point clouds~\cite{qin2023dexpoint, chen2022inhand}, and kinematic descriptions like URDF models~\cite{shao2020unigrasp}, which help facilitate transfer between different hand morphologies. However, when it comes to modeling interaction, most existing methods rely on coarse spatial encodings, such as distances between hand and object~\cite{GRAFF:mandikal2021learning, mandikal2022dexvip} or proprioceptive signals~\cite{joshi2020robotic}. These features typically fall short in capturing the nuanced contact dynamics necessary for precise manipulation. To address this, She \etal~\cite{she2022ibs} proposed the Interaction Bisector Surface (IBS) as a geometric descriptor for grasp synthesis. While effective in encoding global interaction surfaces, IBS lacks sensitivity to local surface variation—an essential property for tasks requiring fine contact control, such as in-hand adjustment or compliant manipulation.

Furthermore, for a manipulation policy, the input is required to represent the interaction status of the environment. Though reinforcement learning or imitation learning for complex multi-fingered manipulation has the potential to learn the perception representations for the interaction status and the control policy in an end-to-end manner like many frameworks in computer vision tasks, it is challenged by sparse successful trials (or "positive samples") due to high dimensional action space of multi-fingered hands or the scarcity of large scale of action demonstration data. Therefore, most existing works~\cite{ILAD:wu2022learning,xu2023unidexgrasp,bao2023dexart,chen2022dextransfer, wan2023unidexgrasp++} adopt a two-stage strategy of first learning a feature representation (e.g., encoding object geometry with pre-trained networks such as PointNet~\cite{qi2017pointnet}) and then training a control policy conditioned on this representation.

However, simply encoding global object geometry often fails to capture the fine-grained contact interactions necessary for high-precision dexterous manipulation. We contend that explicitly modeling local surface geometry in conjunction with hand kinematics is critical for informing control policies, especially in dynamic and contact-rich tasks. For example, ManipNet~\cite{zhang2021manipnet} explores this direction by predicting manipulation trajectories from hand-object spatial relations in a supervised setting. Inspired by this insight, we introduce a compact, structured representation tailored for RL, which captures both the global spatial layout and fine local surface geometry. Our goal is to provide general-purpose features that enable generalizable dexterous policy learning across diverse objects, hand morphologies, and manipulation tasks.

%% file: txt/method.tex
\color{black}

\section{Preliminaries and Method Overview}
\subsection{Preliminaries}

We formulate dexterous manipulation as a Markov Decision Process (MDP), defined by the tuple $\mathcal{M} = (\mathcal{S}, \mathcal{A}, \mathcal{T}, \mathcal{R}, \gamma)$. Here, $\mathcal{S} \in \mathbb{R}^{n}$ and $\mathcal{A} \in \mathbb{R}^{m}$ represent the state and action spaces, respectively; $\mathcal{T}:\mathcal{S} \times \mathcal{A} \rightarrow \mathcal{S}$ denotes the state transition dynamics; $\mathcal{R}:\mathcal{S} \times \mathcal{A} \rightarrow \mathbb{R}$ is the reward function measuring task progress; and $\gamma \in (0, 1]$ is the discount factor.

The objective is to learn a policy $\pi(a \vert s)$ that defines a distribution over actions $a$ given a state $s$, maximizing the expected cumulative discounted reward $ \sum_{t=0}^{\infty} \gamma^{t} r_t$, where $r_t \in \mathcal{R}$. We adopt deep reinforcement learning (DRL) to optimize this policy. The key value functions in this context are defined as:

\begin{equation}
\label{V_pi_Equ}
V^{\pi}(s) = \mathbb{E}_{\pi} \left[ \sum_{k=0}^{\infty} \gamma^k R_{t+k+1} \,\middle|\, S_t = s \right],
\end{equation}

\begin{equation}
\label{Q_pi_Equ}
Q^{\pi}(s, a) = \mathbb{E}_{\pi} \left[ \sum_{k=0}^{\infty} \gamma^k R_{t+k+1} \,\middle|\, S_t = s, A_t = a \right],
\end{equation}

\begin{equation}
\label{A_pi_Equa}
A^{\pi}(s, a) = Q^{\pi}(s, a) - V^{\pi}(s).
\end{equation}

We parameterize the policy as a neural network $\pi_\theta$ and define its performance as the expected return:

\begin{equation}
\label{J_pi_theta_Equ}
J(\pi_\theta) = \mathbb{E}_{\tau \sim \pi_\theta(\tau)} \left[ \sum_{t=0}^{\infty} \gamma^t R_t \right],
\end{equation}
where $\tau$ denotes a trajectory sampled from the policy. The policy parameters $\theta$ are optimized via gradient ascent on $J(\pi_\theta)$, using the policy gradient:

\begin{equation}
\label{nabla_J_pi_Equ}
\nabla_\theta J(\pi_\theta) = \mathbb{E}_{\tau \sim \pi_\theta(\tau)} 
\left[ \nabla_\theta \log \pi_\theta(a_t \vert s_t) A^{\pi_\theta}(s_t, a_t) \right].
\end{equation}

\subsection{Method Overview}

\begin{figure*}[t]
    \centering
    \includegraphics[width=0.8\textwidth]{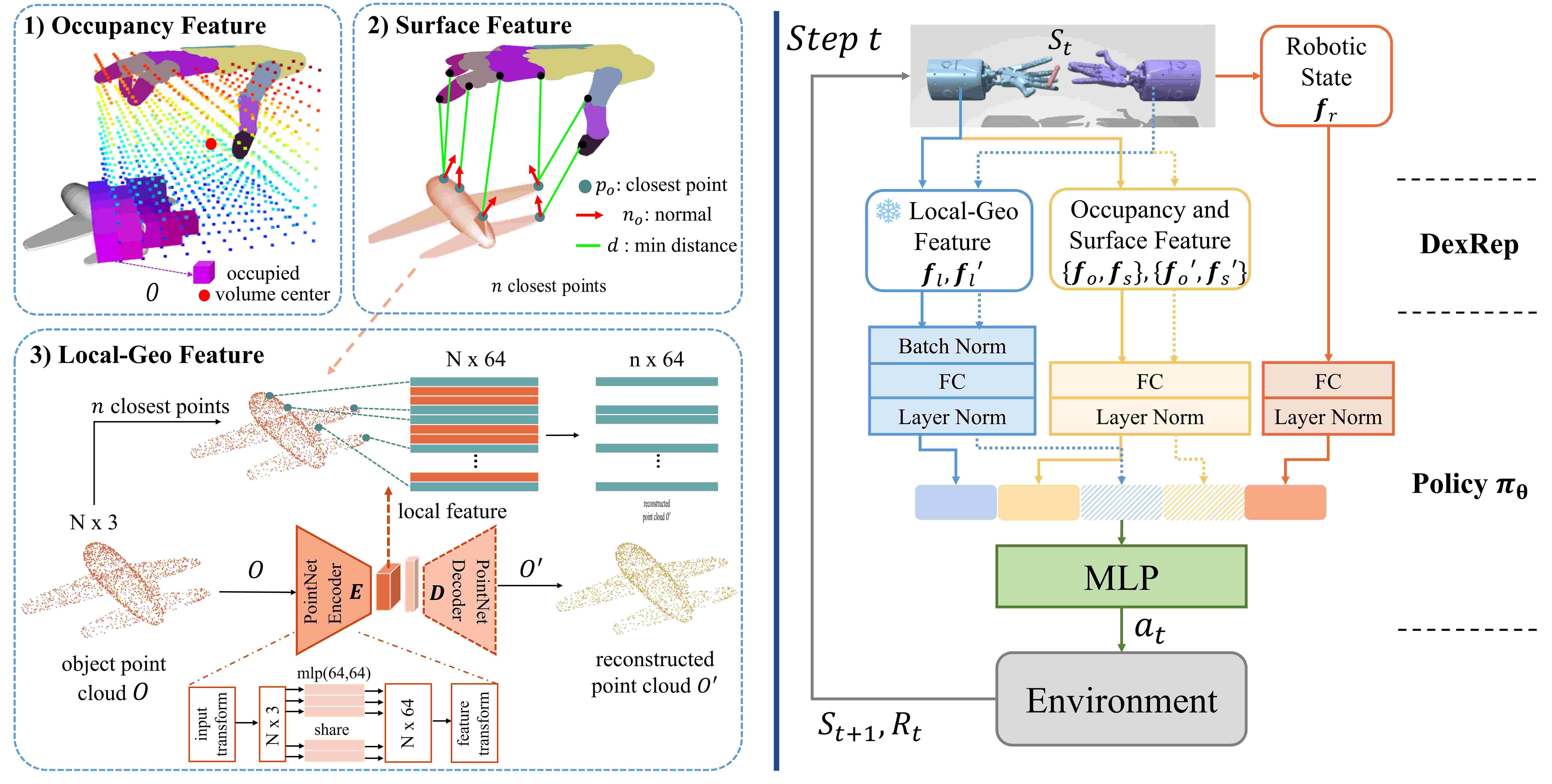}
    \caption{\textbf{\rep and its integration into dexterous manipulation learning.} \textbf{Left:} Visualization of the three components of \rep—Occupancy, Surface, and Local-Geo features—each encoding a different aspect of the hand-object interaction. \textbf{Right:} Policy learning framework with \rep as input. The dashed box denotes the representation for the second hand in bimanual settings, which is omitted in single-hand scenarios.}
    \label{fig:pipeline}
    \vskip -0.5cm
\end{figure*}

Given the task specification, initial hand configuration, and target manipulation goal (e.g., desired object pose), our objective is to generate a sequence of actions that successfully completes the manipulation. As illustrated in Fig.~\ref{fig:pipeline}, our approach begins by extracting an informative hand-object interaction feature $f$ based on the current observation. 
% This feature captures the spatial relationship, contact-relevant geometry, and relative configuration between the robotic hand and the object.
The complete state input to the policy is composed of this interaction feature $f$ and the proprioceptive state of the robot $f_{\text{prop}}$, such as joint angles and velocities. The policy $\pi_\theta$ then predicts the next action $a = \pi_\theta(s)$, where $s = \{f, f_{\text{prop}}\} \in \mathcal{S}$. The environment transitions to the next state via the dynamics $\mathcal{T}$, and the process iterates until task success or failure. Our method is evaluated across multiple dexterous manipulation tasks and demonstrates generalizable performance without requiring task-specific modifications.

% \vspace{0.3em}
The remainder of this paper is organized as follows:  
Sec.~\ref{Section DexRep} introduces our proposed hand-object interaction representation, \rep; 
Sec.~\ref{Section Learning} describes how we train manipulation policies using \rep for different tasks;  
Sec.~\ref{Section DiffTasks Efficacy} presents comparative results across multiple manipulation tasks; 
Sec.~\ref{Sec DexRep Exp} analyzes the characteristics and generalization performance of \rep;  
Sec.~\ref{Sec real world} demonstrates the real-world deployment and robustness of our approach.

\section{\rep: Dexterous Representation} \label{Section DexRep}

In this section, we introduce our key contribution: DexRep, a hand-object interaction representation for learning dexterous manipulation tasks in robotics. \rep extracts features $f$ that provide the policy $\pi$ with rich spatial and geometric cues, enabling effective perception of hand-object configurations and promoting robust generalization across tasks and object geometries. \rep comprises three complementary components: the Occupancy Feature $f_o$, the Surface Feature $f_s$, and the Local-Geo Feature $f_l$.

\subsubsection{Occupancy Feature $f_o$}

The Occupancy Feature captures coarse global geometry by encoding the voxel occupancy within a 3D volume anchored to the palm of the robotic hand. It captures global shape information via a coarse occupancy volume instead of point clouds carrying detailed geometry.
% for high-level generalization.  

% This abstraction offers a simplified yet effective spatial representation for collision avoidance and approach planning.

The volume is defined as a $10 \times 10 \times 10$ voxel grid, where each voxel has an edge length $l_v$, chosen based on the required task-level precision. The volume is anchored to a point slightly inside the palm, offset perpendicularly from the root joint of the middle finger, as this region is most relevant for pre-grasp and contact. The volume rigidly follows the root joint’s position and orientation during manipulation. We define $f_o \in \{0,1\}^{1000}$, where each element $f_o^i$ represents whether the $i$-th voxel is occupied by any point from the object point cloud:
\begin{equation}
\label{OccFea_Equ}
f_o^{i} = \begin{cases}
1, & \text{if occupied,} \\
0, & \text{otherwise.}
\end{cases}
\end{equation}

% This feature encodes the coarse local shape of the object near the hand, providing simplified geometric cues for robust object localization without requiring fine global details. Using a lower-resolution grid keeps the representation simple and more generalizable to diverse objects. Increasing the grid resolution reveals finer surface details but comes at the cost of higher computational overhead and may reduce generalization capability and robustness to noisy observations.

This compact binary encoding facilitates robust object localization and generalization by abstracting away fine-grained shape details. This abstraction offers a simplified yet effective spatial representation for approaching and hand pose pre-shaping. Higher-resolution grids can capture more details, but at the cost of increased computational overhead and reduced generalization to object variation and robustness to input noise.

\subsubsection{Surface Feature $f_s$}

To support precise finger control, we introduce the Surface Feature, which encodes spatial distances and local surface normals between key hand joints and the object surface. Let $n$ denote the number of sampled hand keypoints (e.g., fingertips and joints), then $f_s \in \mathbb{R}^{4n}$ is defined as:

\begin{equation}
\label{SurFea_Equ}
f_s^{j} = \left\{ \min\left( \| p_h^j - p_o^j \|, \sigma_{\max} \right),\; n_o^j \right\},
\end{equation}
where $p_h^j$ is the $j$-th keypoint on the hand, $p_o^j$ is its closest point on the object surface, and $n_o^j$ is the corresponding surface normal. The maximum distance threshold $\sigma_{\max}$ is set based on the expected range of hand-object proximity for each task.

This feature captures two critical cues: (\romannumeral1) the proximity between each finger joint and nearby object surfaces, and (\romannumeral2) the orientation of those surfaces via normals. These cues allow the policy to infer where and how to approach the object for stable and purposeful contact.

\subsubsection{Local-Geo Feature $f_l$}

While surface normals provide directional cues, they do not encode rich geometric structures such as curvature, thickness, or local symmetry, which are essential for fine manipulation. To address this, we propose the Local-Geo Feature $f_l$, which augments the Surface Feature with learned geometric descriptors extracted via a pretrained PointNet encoder.

We adopt a two-stage pipeline: In the first stage, we train a PointNet autoencoder to reconstruct object point clouds from ShapeNet55~\cite{chang2015shapenet}, using a training loss:

\begin{equation}
L_{\text{ae}} = \alpha \cdot L_{\text{cls}} + \beta \cdot L_{\text{chamfer}} + \gamma \cdot L_{\text{emd}},
\end{equation}
where $L_{\text{cls}}$ is the classification loss, $L_{\text{chamfer}}$ measures average point-wise distance~\cite{achlioptas2018learning}, and $L_{\text{emd}}$ is the Earth Mover’s Distance~\cite{fan2017point}. We set $\alpha = 2$, $\beta = 0.1$, and $\gamma = 100$ to balance shape reconstruction and feature discrimination.

After training, we retain only the encoder to extract per-point descriptors. For each hand keypoint, we retrieve the descriptor of its closest point on the object surface, yielding $f_l \in \mathbb{R}^{n \times 64}$. This feature provides compact yet expressive local geometry, facilitating nuanced contact reasoning in manipulation.

\subsubsection{Discussion}

\rep provides a unified and flexible representation that balances coarse global structure and fine local detail. The voxel-based Occupancy Feature is simple to compute and robust to noise, making it suitable for coarse alignment and collision avoidance. The Surface Feature captures critical contact geometry through relative distances and normals. The Local-Geo Feature offers high-capacity geometric descriptors, enhancing performance in tasks requiring fine contact adaptation.

We choose voxel grids for global shape abstraction due to their computational simplicity and discretization properties. While denser sampling on the hand surface could enrich spatial resolution, it would also increase computational complexity. Surface normals offer an interpretable and low-dimensional encoding, whereas PointNet descriptors serve as a more expressive alternative for complex geometries.

Together, these features enable DexRep to serve as a general-purpose, task-agnostic interaction representation that supports robust learning and transfer in dexterous manipulation.

\section{Learning Manipulation Policy with \rep}  \label{Section Learning}

In this section, we describe how we incorporate \rep into reinforcement learning (RL) to train manipulation policies $\pi_\theta$ for three dexterous tasks: grasping, in-hand reorientation, and handover. We first present the network architecture that integrates \rep as a policy input. We then define the reward functions and action spaces tailored to each task. Finally, we describe how we train the policy network $\pi_\theta$ using RL with or without demonstrations depending on manipulation tasks.

\subsection{Embedding \rep into RL Policy $\pi_\theta$} \label{Sec Learn Policy}

To support both single- and dual-hand scenarios, we adopt a modular network design. For each hand, we compute \rep features—including the Occupancy ($f_o$), Surface ($f_s$), and Local-Geo ($f_l$) components. When dual hands are used, the features from the second hand are denoted by a prime symbol (e.g., $f_o'$). We also include the proprioceptive state $f_{prop}$, which encodes the joint angles, velocities, and other robot-specific information.

The network architecture processes each feature component individually. Occupancy Feature and Surface Feature $\{f_o, f_s\}$ (and $\{f_o', f_s'\}$ if present) as well as $f_{prop}$ are passed through independent fully connected (FC) layers with layer normalization. Due to its high magnitude variability, the Local-Geo feature $f_l$ undergoes batch normalization before being processed by a dedicated FC layer and layer normalization. The resulting feature embeddings are concatenated and input to a multi-layer perceptron (MLP), which outputs the action $a$. The full network architecture is illustrated in Fig.~\ref{fig:pipeline} (right).

\begin{figure*}[!t]
    \centering
    \begin{minipage}{0.48\linewidth}
        \includegraphics[width=\linewidth]{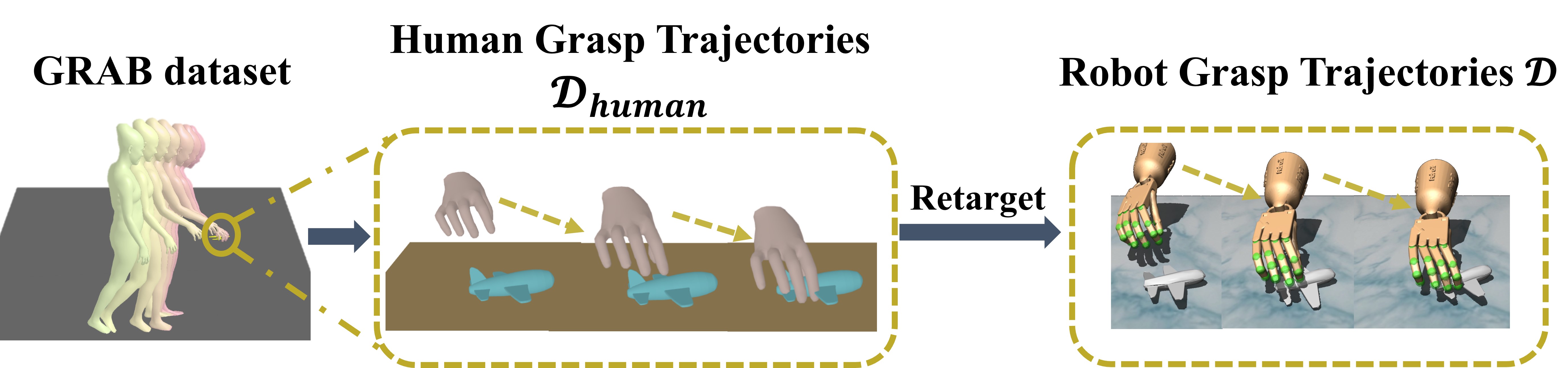}
        \caption{\textbf{Demonstration acquisition for behavior cloning.} In the grasping task, we obtain human demonstration data $\mathcal{D}_{\text{human}}$ from the GRAB dataset~\cite{taheri2020grab} and retarget it to Adroit hand~\cite{kumar2013adroithand} to generate robot demonstration data $\mathcal{D}$ for subsequent BC initialization of the policy $\pi_\theta$.}
        \label{fig:demo_generate}
    \end{minipage}\hfill
    \begin{minipage}{0.49\linewidth}
        \includegraphics[width=\linewidth]{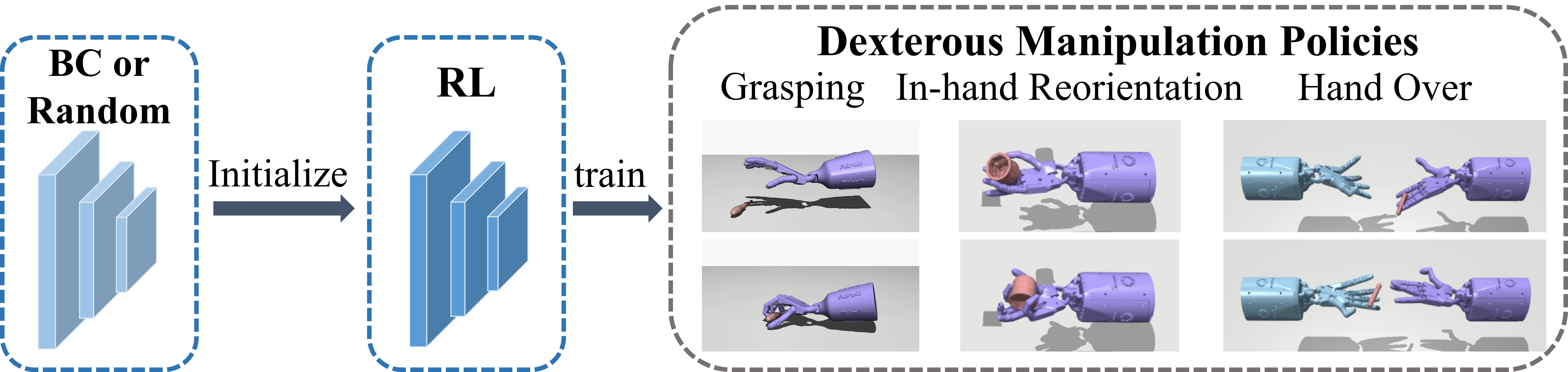}
        \vspace{-0.5cm}
        
        \caption{\textbf{Our pipeline to learn manipulation policy with \rep.} For the grasping task, we first pre-train the policy using BC and then fine-tune it through RL. For in-hand reorientation and handover tasks, we start with a randomly initialized policy and learn the strategy from scratch using RL.}
        \label{fig:learning_pipeline}
    \end{minipage}

\vskip -0.5cm
\end{figure*}

\subsection{Reward Function $\mathcal{R}$}    \label{Sec Reward}

\subsubsection{Grasping Task}
The total reward is composed of four components:
\begin{equation}
r_{\text{grasp}} = r_{\text{approach}} + r_{\text{lift}} + r_{\text{pen}} + r_{\text{target}}.
\end{equation}

\textbf{Approach Reward:} encourages the hand to move closer to the object:
\begin{equation}
r_{\text{approach}} = 0.1 \left( \frac{1}{10 \cdot d_{f\rightarrow o} + 0.25} - 1 \right),
\end{equation}
where \(d_{f\rightarrow o}\) is the sum of distances from all fingertips to the object surface.

\textbf{Lift Reward:} encourages the object to be lifted above the table:
\begin{equation}
r_{\text{lift}} = 
\begin{cases}
1, & \text{if } h > 0.02~m, \\
0, & \text{otherwise},
\end{cases}
\end{equation}
where \(h\) is the object’s height above the table.

\textbf{Penetration Penalty:} discourages penetration into the object or table:
\begin{equation}
r_{\text{pen}} = \max\left( 0.01 \left( 1 - e^{-(d_{\text{table}} + d_{\text{obj}})} \right), -100 \right),
\end{equation}
where \(d_{\text{table}}\) and \(d_{\text{obj}}\) are signed penetration depths.

\textbf{Target Reward:} encourages object delivery to the target location without penetration:
\begin{equation}
r_{\text{target}} =
\begin{cases}
10, & \|p_{\text{obj}} - p_{\text{tar}}\| < 0.1~m \text{ and } r_{\text{pen}} > -30, \\
20, & \|p_{\text{obj}} - p_{\text{tar}}\| < 0.05~m \text{ and } r_{\text{pen}} > -30, \\
0, & \text{otherwise}.
\end{cases}
\end{equation}

\subsubsection{In-Hand Reorientation Task}
The reward consists of four components:
\begin{equation}
r_{\text{rot}} = r_{\text{dis}} + r_{\text{orient}} + r_{\text{act}} + r_{\text{suc}}.
\end{equation}

\textbf{Distance Reward:} penalizes deviation from the target position:
\begin{equation}
r_{\text{dis}} = -10 \cdot d_{o\rightarrow t},
\end{equation}
where \(d_{o\rightarrow t}\) is the Euclidean distance between the object and the target.

\textbf{Orientation Reward:} encourages rotation alignment:
\begin{equation}
r_{\text{orient}} = \frac{1}{|d_{\text{rot}}| + 0.1},
\end{equation}
where \(d_{\text{rot}} = 2 \arcsin(\min(\|(\Delta x_q, \Delta y_q, \Delta z_q)\|_2, 1.0))\), derived from quaternion difference.

\textbf{Action Penalty:} discourages large joint movements:
\begin{equation}
r_{\text{act}} = -0.0002 \sum_{i=1}^{n_j} a_i^2,
\end{equation}
where \(a_i\) is the action on joint \(i\), and \(n_j\) is the number of joints.

\textbf{Success Reward:} provides a bonus when rotation is sufficiently accurate:
\begin{equation}
r_{\text{suc}} = 
\begin{cases}
250, & \text{if } d_{\text{rot}} \leq 0.1~\text{rad}, \\
0, & \text{otherwise}.
\end{cases}
\end{equation}

\subsubsection{Handover Task}
The handover reward encourages accurate object delivery via:
\begin{equation}
r_{\text{over}} = 
\begin{cases}
e^{-10 \cdot d_{o\rightarrow t}} + 250, & \text{if } d_{o\rightarrow t} \leq 0.02~m, \\
e^{-10 \cdot d_{o\rightarrow t}}, & \text{otherwise},
\end{cases}
\end{equation}
where \(d_{o\rightarrow t}\) is the Euclidean distance between the object and the target pose.

\subsection{Action Space $\mathcal{A}$} \label{sec action}

We define $\mathcal{A}$ as a continuous space over target joint positions. For grasping, the action is $a = \{p_g, \theta\}$ where \(p_g \in \mathbb{R}^6\) is the global pose of the Adroit hand and \(\theta \in \mathbb{R}^{24}\) is the set of joint angles. For the other two tasks using the Shadow Hand, \(a = \theta \in \mathbb{R}^{20}\).

Each action \(a_{\text{target}}\) is executed using a low-level PD controller:
\[
\tau = K_p (a_{\text{target}} - a_{\text{current}}) - K_d \, \dot{a}_{\text{current}},
\]
where \(a_{\text{current}}\) and \(\dot{a}_{\text{current}}\) denote the current joint positions and velocities, respectively, and \(K_p\) and \(K_d\) are the proportional and derivative gains. The gains \(K_p\) and \(K_d\) are hand-specific and we follow \cite{DAPG:rajeswaran2017learning, chen2022issacManipulation} to set the values.

\subsection{Learning Manipulation Policies via $\nabla_\theta J(\pi_\theta)$}

\subsubsection{Learning from Demonstrations for Grasping}

Incorporating human demonstrations has proven effective in improving sample efficiency and promoting safe, human-like robotic behaviors~\cite{DAPG:rajeswaran2017learning, ILAD:wu2022learning, qin2022dexmv}. Following this line of work, we adopt a two-stage learning strategy for the grasping task: we first pretrain the policy $\pi_\theta$ using Behavior Cloning (BC) from expert demonstrations, and then fine-tune it using reinforcement learning (RL). The overall pipeline is illustrated in Fig.~\ref{fig:demo_generate} and Fig.~\ref{fig:learning_pipeline}.

To acquire demonstrations, we utilize a retargeting-based pipeline to convert human grasp trajectories from GRAB~\cite{taheri2020grab} into robot-executable demonstrations (Fig.~\ref{fig:demo_generate}). GRAB provides whole-body motion sequences of 10 subjects interacting with 51 household objects. We extract the hand-object interaction segment from each trajectory using the MANO hand model~\cite{MANO:SIGGRAPHASIA:2017}, specifically cropping frames from when the hand is within $10~cm$ of the object until the object is lifted by $4~cm$. All demonstrations are normalized to an object-centric coordinate frame by translating the object to the origin and applying the same transformation to the hand.

\begin{figure}[!h]
    \centering
    \vspace{-0.5cm}
    \includegraphics[width=0.7\linewidth]{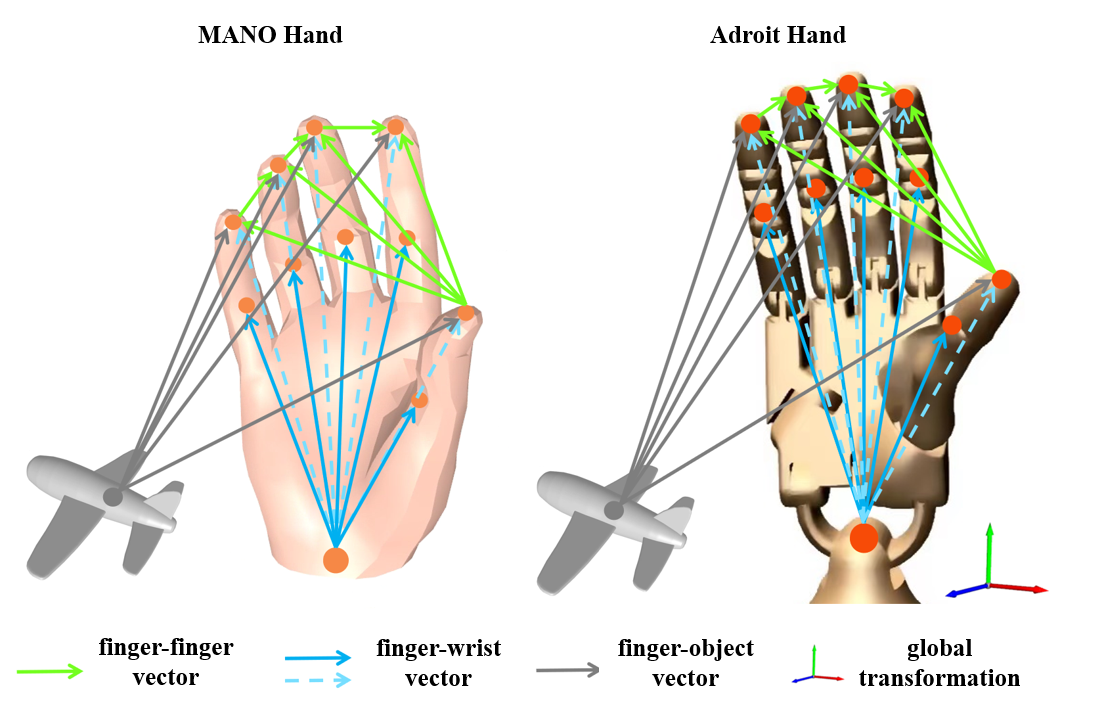}
    \vspace{-0.2cm}
    \caption{\textbf{Retargeting process.} Key vectors—finger-to-finger, finger-to-wrist, and finger-to-object—are computed from both human (MANO) and robotic (Adroit) hands and optimized to align hand postures.}
    \label{fig:retarget}
\end{figure}

Let $\mathcal{D}_{\text{Human}} = \{ d_0, \dots, d_N \}$ denote the set of human demonstrations, where each trajectory $d_n = \{ (g_t, q_t, o_t) \}_{t=0}^{T}$ contains global hand poses $g_t$, joint angles $q_t$, and object poses $o_t$ at each timestep $t$. We aim to map these to the Adroit hand~\cite{kumar2013adroithand} via retargeting.

Following DexPilot~\cite{handa2020dexpilot}, we independently optimize each frame's joint configuration via non-linear optimization. The objective minimizes discrepancies between key postural vectors of the human and robot hands:
\begin{equation}
\label{retarget}
\min_{q_{t}^{\mathbf{R}}, R_g} \sum_{i=0}^{I} \left\| \mathbf{v}_{i}^{\mathbf{R}}(R_g, q_{t}^{\mathbf{R}}) - k_i \, \mathbf{v}_{i}^{\mathbf{H}}(q_{t}^{\mathbf{H}}) \right\|^2,
\end{equation}
where $I=3$ represents three kinds of key vectors (finger-to-finger vectors, finger-to-wrist vectors, finger-to-object vectors) of the robotic hand $\mathbf{v}_{\mathbf{i}}^{\mathbf{R}}$ and the human hand $\mathbf{v}_{\mathbf{i}}^{\mathbf{H}}$,  computed by the joint angles $q_{t}^{\mathbf{R}}$ and $q_{t}^{\mathbf{H}}$ through forward kinematics. The three types of key vector are shown in Fig.~\ref{fig:retarget}. Here, \(k_i\) is the scale ratio for the \(i^{th}\) type of key vectors between the Adroit Hand~\cite{kumar2013adroithand} and the MANO Hand~\cite{MANO:SIGGRAPHASIA:2017}, and \(R_g\) denotes the global translation and rotation of the robotic arm attached to the hand. Including \(R_g\) in optimization helps reduce positional error caused by structural differences between human and robotic hands.

After optimization, we convert the retargeted joint trajectories into executable actions in MuJoCo~\cite{todorov2012mujoco}, optionally applying correlated sampling~\cite{chen2022dextransfer} to reduce object drop during lifting. This yields a robot demonstration dataset $\mathcal{D} = \{(s, a)\}$ for subsequent BC training.

The policy is then initialized using Behavior Cloning:
\begin{equation}
    L_{\text{BC}} = \frac{1}{|\mathcal{D}|} \sum_{(s, a) \in \mathcal{D}} \left\| \pi_\theta(s) - a \right\|^2,
    \label{BC_grad_Equ}
\end{equation}
where $|\mathcal{D}|$ is the number of state-action pairs in the dataset and  $\pi_\theta$ is the policy network parameterized by $\theta$. While BC offers a strong initialization, it is susceptible to covariate shift between the demonstration and learned policy distributions.

To mitigate this, we adopt the DAPG algorithm~\cite{DAPG:rajeswaran2017learning}, which integrates demonstrations into RL fine-tuning. Specifically, we augment the standard policy gradient (Eq.~\ref{nabla_J_pi_Equ}) with an additional term based on demonstration data:
\begin{align}
    \nabla_\theta J(\theta)_{\text{FT}} 
    &= \mathbb{E}_{(s,a)\sim\rho^\pi}\left[\nabla_\theta \ln \pi_\theta(a|s) A^\pi(s,a)\right] \\
    + &\mathbb{E}_{(s,a)\sim \mathcal{D}}\left[
        \nabla_\theta \ln \pi_\theta(a|s) \lambda_0 \lambda_1^k \max_{(s',a')\in\rho^\pi} A^\pi(s',a') \right], \nonumber
\end{align}
where $\rho^\pi$ is the current policy rollout, $A^\pi$ is the advantage function (computed via GAE~\cite{GAE:schulman2015high}), $(\lambda_0, \lambda_1) = (0.1, 0.95)$ control the decaying contribution of demonstrations, and $\max _{(s', a') \in \rho_{\pi}} A^{\pi_{\theta}}(s', a')$=1.

% \vspace{0.2cm}
\subsubsection{Learning from Scratch for Other Tasks}

Following prior work~\cite{chen2022issacManipulation}, we train policies for in-hand reorientation and handover tasks using PPO~\cite{schulman2017ppo} from scratch in simulation, leveraging massive parallelization via Isaac Gym~\cite{makoviychuk2021isaac}. DexRep is used as the primary state representation.

PPO maintains two networks: the actor $\pi_\theta(a | s)$ and the critic $V_\phi(s)$. The actor is updated via:
\begin{align}
    \nabla_\theta J(\theta)_{\text{PPO}} =
    \mathbb{E}_{(s,a)\sim\rho^{\pi_{\theta_\mathrm{old}}}}
    \big[ &\nabla_\theta r_t(\theta) A^{\pi_\theta}(s,a) \\
    & - \beta \nabla_\theta \mathrm{KL}\big(\pi_{\theta_{\mathrm{old}}} \| \pi_\theta\big) \big], \nonumber
\end{align}
where $r_t(\theta) = \frac{\pi_{\theta}(a\vert s)}{\pi_{\theta_{\text{old}}}(a \vert s)}$ is the probability ratio, $\beta$ determines the weight of the KL divergence term in the objective, and $A^{\pi_{\theta}}$ is the advantage function and is calculated using GAE~\cite{GAE:schulman2015high}.

The critic is updated to minimize the squared error between the predicted and target returns:
\begin{equation}
    \nabla_\phi J(\phi)_{\text{PPO}} =
    \mathbb{E}_{(s,a)\sim\rho^{\pi_{\theta_\mathrm{old}}}}
    \left[ 2 \left( V_\phi(s) - G \right) \nabla_\phi V_\phi(s) \right],
\end{equation}
where $G = A + V_{\phi_{\text{old}}}(s)$ is the bootstrapped return using the old critic network.

This learning setup allows us to effectively train policies for complex dexterous tasks without relying on demonstrations.

%% file: txt/exper_result.tex
\section{Effectiveness of DexRep Across Tasks}  \label{Section DiffTasks Efficacy}

To comprehensively evaluate the robustness and generalization capability of \rep across diverse dexterous manipulation scenarios, we benchmark its performance in three representative tasks: grasping, in-hand reorientation, and bimanual handover. For each task, we compare DexRep against state-of-the-art baselines in terms of success rate, training efficiency, and generalization to unseen objects. \lqtt{We conduct each simulation experiment with 10 random seeds in the paper, and report the mean and standard deviation of the seeds. In all tables, ``±'' denotes the standard deviation (SD) of the reported success rates.}

\begin{figure*}[!t]
    \centering
    \begin{minipage}{\linewidth}
    \centering
        \includegraphics[width=0.85\linewidth]{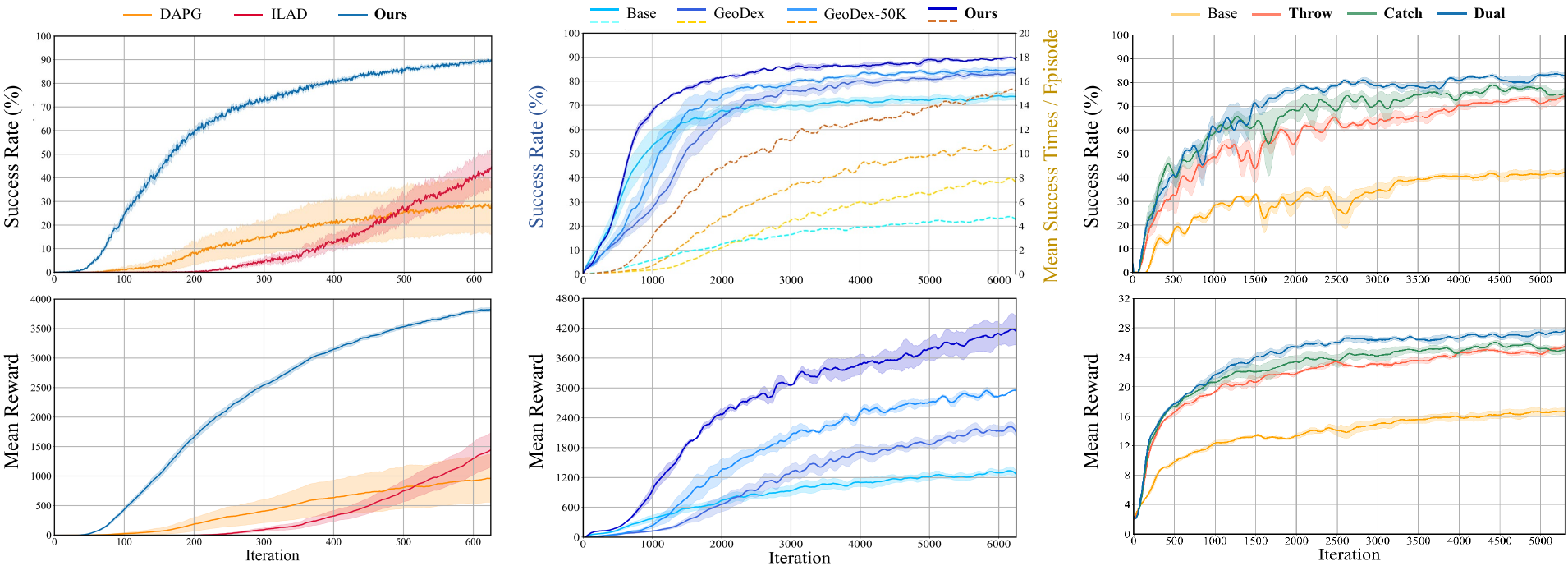}

        \caption{\lqtt{\textbf{Success rate (or mean success times per episode) and mean reward of our method and baselines in different tasks during training (left: Grasping; middle: In-Hand Reorientation; right: Handover).} The x-axis represents the training iterations. The y-axis indicates the success rate (\%), mean success times per episode \textbf{(dashed lines)}, or mean reward, and the shaded area represents the standard deviation.} }
        \label{fig:tasks_train}
    \end{minipage}

    \begin{minipage}{\linewidth}
        \begin{table}[H]
        \centering
       
                \begin{threeparttable}
  \caption{\textbf{Success rate (\%) of our method and baselines for grasping in MuJoCo and Isaac Gym.} }
    \begin{tabular}{l|ccc|cccccc}
 
    \toprule[0.5mm]
    Simulator & \multicolumn{3}{c|}{MuJoCo} & \multicolumn{6}{c}{Isaac Gym} \\
    \midrule
    Methods & DAPG~\cite{DAPG:rajeswaran2017learning}   & ILAD~\cite{ILAD:wu2022learning}   & \textbf{Ours}   & DAPG~\cite{DAPG:rajeswaran2017learning}   & ILAD~\cite{ILAD:wu2022learning}   & IBS~\cite{she2022ibs}    & UniDexGrasp~\cite{xu2023unidexgrasp} & UniDexGrasp++~\cite{wan2023unidexgrasp++} & \textbf{Ours} \\
    \midrule
    Seen   & \lqtt{36.3±8.2} & \lqtt{64.8±3.3} & \lqtt{\textbf{96.5±0.9}} & 21.0   & 32.0   & 57.0   & 79.0   & 85.4   & \textbf{\lqtt{96.5}}  \\
    Unseen & \lqtt{21.5±12.3} & \lqtt{22.6±3.1} & \lqtt{\textbf{88.1±2.0}} & 12.5   & 24.5   & 54.0   & 72.5   & 78.2   & \textbf{\lqtt{96.4}}  \\
   
     \bottomrule[0.5mm]
    \end{tabular}%
  \label{tab:grasp-test}%
      \begin{tablenotes}
        \footnotesize 
        \item Notes: Some simulation experiment settings, like training and testing objects and the reward functions in MuJoCo and Isaac Gym, also differ. See Sec. \ref{effectiveness_grasping} for more details. The results of DAPG, ILAD, and IBS in Isaac Gym are sourced from UniDexGrasp~\cite{xu2023unidexgrasp}.
    \end{tablenotes}
    \end{threeparttable}
            
        \end{table}
    \end{minipage}
    \begin{minipage}{\linewidth}
                % Table generated by Excel2LaTeX from sheet 'Sheet2'
        \begin{table}[H]
          \centering
          \caption{\textbf{Success rate (\%) of our method and baselines for In-Hand Reorientation and Handover.} ``GeoDex-50k" is a variant of GeoDex, which is pretrained with the same 50K objects as our method.}
            \begin{tabular}{l|cccc|cccc}
            % \toprule
            \toprule[0.5mm]
            Tasks  & \multicolumn{4}{c|}{In-Hand Reorientation } & \multicolumn{4}{c}{Handover} \\
            \midrule
            Methods & Base~\cite{chen2022issacManipulation}   & GeoDex~\cite{huang2021generalization} & {GeoDex-50k} & \textbf{Ours}   & Base~\cite{chen2022issacManipulation}   & Ours-Throw & Ours-Catch & \textbf{Ours-Dual} \\
            \midrule

            Seen	& \lqtt{66.5±3.9}	& \lqtt{87.5±0.8}	& \lqtt{88.8±0.1}	& \lqtt{\textbf{91.1±1.2}}	& \lqtt{43.1±3.2}	& \lqtt{62.3±4.2}	& \lqtt{77.3±4.7}	& \lqtt{\textbf{81.5±0.1}} \\
            Unseen	& \lqtt{50.1±6.2}	& \lqtt{76.3±3.0}	& \lqtt{81.1±1.5} & \lqtt{\textbf{86.0±3.9}}	& \lqtt{36.8±2.9}	& \lqtt{60.0±1.2}	& \lqtt{72.1±3.8}	& \lqtt{\textbf{77.3±1.9}} \\

            % \bottomrule
             \bottomrule[0.5mm]
            \end{tabular}%
          \label{tab:inhand&handover-test}%
        \end{table}%

    \end{minipage}
    \vspace{-0.5cm}
\end{figure*}

\subsection{Efficacy of DexRep in Grasping}

We first evaluate the performance of \rep in dexterous grasping tasks, focusing on its training efficiency and generalization to unseen objects.

\subsubsection{Experimental Setup}  \label{Sec Grasp Exp Settings}

Experiments are conducted in the MuJoCo~\cite{todorov2012mujoco} simulator using the 30-DoF AdroitHand~\cite{kumar2013adroithand} (Fig.~\ref{fig:teaser}, left). At each episode’s start, the object is placed near the origin with small perturbations: translation in $x \sim [-0.05, 0.05]$\,m, $y \sim [-0.05, 0]$\,m, and a random in-plane rotation in $[-\pi, \pi]$. Object mass and friction are set to $0.5$\,kg and $1.0$. The hand is initialized to the average pre-grasp pose of all retargeted demonstrations.

\textit{\textbf{Baselines.}}  
We compare \rep against two widely used hand-object representations in dexterous RL:  
(1)~\textbf{DAPG}~\cite{DAPG:rajeswaran2017learning}: uses relative hand-object distances;  
(2)~\textbf{ILAD}~\cite{ILAD:wu2022learning}: uses PointNet~\cite{qi2017pointnet} to encode object shape features.  
We reimplement both in the same setup for fair comparison.

\textit{\textbf{Datasets.}}  
Training and evaluation follow a standard object split. We train using 40 objects from GRAB~\cite{taheri2020grab} and evaluate both on these (seen) and the full set of DexGraspNet~\cite{wang2023DexGraspNet} (5,355 unseen shapes). 

\textit{\textbf{Network Architecture.}}  
The input to the policy includes the robot state $f_{prop}$ and the three DexRep features. Input sizes are 105 ($f_{prop}$), 1088 ($f_o + f_s$), and 1408 ($f_l$). Each input is passed through a single-layer encoder of size 128. The policy itself is a 3-layer MLP with hidden dimensions $[512, 128]$, followed by ReLU activations. The output action $\mathbf{a} \in \mathbb{R}^{30}$ includes 6-DoF end-effector pose and 24 joint commands.

\textit{\textbf{Training Details.}}  
DexRep is computed with a voxel size $l_v = 0.02$\,m and surface feature perception range $\sigma_{max} = 0.2$\,m (Based on ablation in Sec.~\ref{Sec Ablation features parameter}).
Behavior Cloning (BC) is used to warm-start policies using 40-object demonstrations, optimized with Adam (lr=$1\times10^{-5}$, batch=64, 150 epochs). BC converges in $\sim$6 minutes.  
Subsequently, we train the policy with DAPG for 600 iterations using \lqtt{10} seeds. Each iteration generates 10 trajectories per object, each of 200 steps. RL training takes approximately 42 GPU hours on an NVIDIA RTX 3090 GPU.

\textit{\textbf{Evaluation Metric.}}  
Following UniDexGrasp~\cite{xu2023unidexgrasp}, a grasp is successful if the object is lifted to $30\pm5$\, cm above the table by the end of the episode. The final evaluation is over 100 trajectories per object. 

\subsubsection{Quantitative Results and Analysis} \label{effectiveness_grasping}

As shown in Fig.~\ref{fig:tasks_train} (left), \textbf{\rep} achieves faster convergence and higher final success rate than DAPG and ILAD. On 5,355 unseen DexGraspNet objects, \textbf{\rep} achieves \lqtt{{88.1\%}} success, outperforming \textbf{ILAD} and \textbf{DAPG} by about \lqtt{\textbf{66\%}} (Table \ref{tab:grasp-test}, left). This demonstrates that DexRep, trained on just 40 objects, generalizes remarkably well to unseen categories with large geometric variation. To compare with UniDexGrasp~\cite{xu2023unidexgrasp} and UniDexGrasp++~\cite{wan2023unidexgrasp++}, we also train \rep in Isaac Gym using their protocols. As shown in Table~\ref{tab:grasp-test} (right), \textbf{\rep} achieves \textbf{$\sim$96\%} success on both seen and unseen objects, exceeding \textbf{UniDexGrasp++} by \lqtt{\textbf{11.1\%}} (seen) and \lqtt{\textbf{18.2\%}} (unseen), despite using a smaller training set. This confirms the portability and robustness of \rep across different RL frameworks and simulation engines.
 
The large performance gap between DexRep and geometry-agnostic (DAPG) or globally encoded (ILAD) methods highlights the importance of structured local interaction modeling. Our results indicate that surface-aware local features are essential not only for fine-grained control but also for generalizable policy learning. Interestingly, the improvement is more prominent in unseen-object tests, implying that DexRep captures task-relevant affordances beyond mere geometry similarity.

\subsection{Efficacy of DexRep in In-Hand Reorientation} \label{Sec DexRep inhand}

We evaluate the effectiveness of \rep in dynamic in-hand reorientation, where the robotic hand must rotate an in-hand object to reach a target orientation with high precision.

\subsubsection{Experimental Setup} \label{Sec InHand Exp Settings}
Following~\cite{chen2022issacManipulation, chen2022inhand}, we conduct experiments in Isaac Gym~\cite{makoviychuk2021isaac} using the 24-DoF Shadow Hand~\cite{shadowrobot_hand} (Fig.~\ref{fig:teaser}, middle). At each episode’s start, the hand is placed at $(0, 0, 0.5)$ in the world frame, with its joint angles perturbed by $\mathcal{N}(0, 0.2)$. The object is initialized at an offset of $(0, -0.39, 0.1)$ from the hand, with additional Gaussian noise $\mathcal{N}(0, 0.01)$. Both initial and target orientations are randomly sampled from $\mathrm{SO}(3)$.

\textit{\textbf{Baselines.}}  
We compare \rep against three baselines trained from scratch using PPO~\cite{schulman2017ppo}:
(1)~\textbf{Base}~\cite{chen2022issacManipulation}: includes full proprioception and object pose/velocity;
(3)~\textbf{GeoDex}~\cite{huang2021generalization}: augments Base with a rotation-sensitive PointNet encoder trained on 85 objects.
(3)~\textbf{GeoDex-50k}~\cite{huang2021generalization}: augments Base with a rotation-sensitive PointNet encoder trained on the same object set (50k objects) as ours.

\begin{figure}[h!]
    \centering
    \vspace{-0.3cm}
    \includegraphics[trim=0 5 5 2, clip, width=0.95\linewidth]{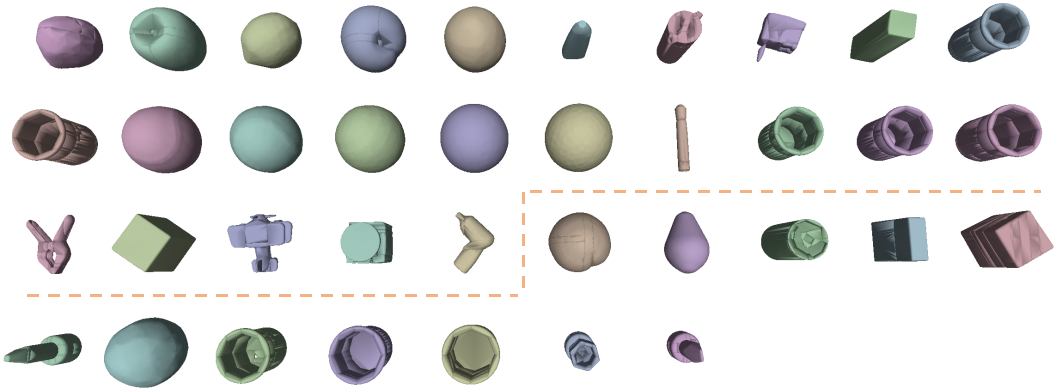}
    \caption{37 Objects used in In-Hand Reorientation. The first 25 objects (above the dotted line) are used in training (seen), while the rest are for evaluation (unseen).}
    \label{fig:reorientation-test_objs}
\end{figure}

\textit{\textbf{Datasets.}}  
We select 25 objects for training and 12 for testing (Fig.~\ref{fig:reorientation-test_objs}) from the YCB dataset~\cite{calli2017ycb}, which is used in \cite{huang2021generalization}.  
For the GeoDex baseline, we pretrain its rotation-sensitive PointNet on 85 objects (YCB~\cite{calli2017ycb} + ContactDB~\cite{brahmbhatt2019contactdb}) and also with a larger set (50k ShapeNet objects~\cite{chang2015shapenet}, the same dataset used in Local-Geo Feature of \rep) for an extended comparison.

\textit{\textbf{Network Architecture.}}  
The input consists of $f_{prop}$ (Base), and the DexRep features $f_o$, $f_s$, and $f_l$, with input sizes 211, 1080, and 1280, respectively. Each is passed through a single-layer encoder of size 128. The policy is an MLP with layers $[512, 512, 128]$ and Elu activations. The output $\mathbf{a} \in \mathbb{R}^{20}$ corresponds to 20 joint control signals.

\textit{\textbf{Training Details.}}  
DexRep is computed with voxel size $l_v = 0.01$\,m and surface feature distance $\sigma_{max} = 0.1$\,m (Based on ablation in Sec .~\ref{Sec Ablation features parameter}). RL training is conducted in 8,000 parallel environments for 5,000 iterations. The environment is reset only once at the beginning. If the object reaches the target position prematurely, the target position will be refreshed to ensure the continuity of the hand and object trajectories within the episode. Training takes $\sim$70 GPU hours on an NVIDIA RTX 3090.

\textit{\textbf{Evaluation Metric.}}  
We evaluate using:  
(1)~\textbf{Success Rate}: percentage of episodes with final orientation error $d_{\text{rot}} < 0.1$ rad;  
(2)~\textbf{Success Times per Episode}: average number of successful reorientations per episode;  
(3)~\textbf{Mean Reward}: average return over the last 200 successful episodes.

\subsubsection{Quantitative Results and Analysis}

Table~\ref{tab:inhand&handover-test} reports the success rates of different methods on both seen and unseen objects. Our method (\textbf{Ours}) achieves the highest performance across both sets, reaching a success rate of \lqtt{{91.1\%}} on seen and \lqtt{{86.0\%}} on unseen objects. Compared to the \textbf{Base} representation~\cite{chen2022issacManipulation}, which only includes hand proprioception and object state (pose and velocity), \rep improves performance by \lqtt{{\textbf{24.6\%}}} on seen and \lqtt{{\textbf{35.9\%}}} on unseen objects, demonstrating its significant advantage in tasks requiring precise control.

GeoDex~\cite{huang2021generalization}, a strong baseline, enhances Base by encoding object geometry using a \emph{rotation-sensitive PointNet}, which learns to capture orientation-aware global features. This design is tailored for in-hand reorientation, where object orientation is central to task success. When trained with the original 85 objects, GeoDex achieves \lqtt{87.5\%} (seen) and \lqtt{76.3\%} (unseen). Our retrained version, \textbf{GeoDex-50k}, which uses the same pretraining dataset as our method, improves results to \lqtt{88.8\%} and \lqtt{81.1\%}. Despite the task-specific design of GeoDex, \rep still outperforms both versions of it by \lqtt{\textbf{3.6\%}} on seen and \lqtt{\textbf{9.7\%}} on unseen objects. This performance gap suggests that beyond capturing object orientation, the ability to encode local surface properties and hand-object spatial interaction—as realized in DexRep—is more beneficial for robust and generalizable skill learning.

While GeoDex leverages global geometry to learn object pose sensitivity, it lacks explicit modeling of physical hand-object interactions, such as local geometry variations and the spatial proximity between hands and the surface of objects. In contrast, \rep incorporates both coarse and fine-grained interaction cues—voxelized global geometry, local features, and proximity-based local interaction fields—providing a richer and more structured representation. The voxelized global geometry encodes the object pose, local features provide abundant information for precise contact reasoning, and the spatial proximity helps the control refinement, especially when adapting to novel object shapes and poses. The consistently better performance on unseen objects further validates the generalization strength of structured local interaction modeling.

\subsection{Efficacy of DexRep in Bimanual Handover}
\label{Sec DexRep Throw}

While previous sections demonstrate the efficacy of \rep in single-hand tasks such as grasping and in-hand reorientation, we now investigate whether \rep can enhance synergy in two-hand coordination tasks by modeling hand-object interactions for both hands. To this end, we evaluate \rep in a bimanual handover scenario, where an object must be transferred smoothly from one hand to another.

\subsubsection{Experimental Setup} \label{Sec Throw Exp Settings}

We construct the handover environment in Isaac Gym~\cite{makoviychuk2021isaac}, following settings from~\cite{chen2022issacManipulation}. As illustrated in Fig.~\ref{fig:teaser} (right), two Shadow Hands are fixed in the scene at positions $(0, 0, 0.5)$ (throwing hand) and $(0, -1, 0.5)$ (catching hand), with opposing orientations. The object is initially placed near the catching hand with small random perturbations. Its target pose is offset from the initial pose by $(0, -0.25, 0)$ with a randomly sampled target orientation from \(\mathrm{SO}(3)\).

\textit{\textbf{Baselines.}}  
We compare against the \textbf{Base} representation~\cite{chen2022issacManipulation}, which encodes the pose, velocity, and proprioceptive state of both hands and the object. We then enhance this with \rep in three configurations: \textbf{{Ours-Throw}} (DexRep for throwing hand), \textbf{{Ours-Catch}} (DexRep for catching hand), and \textbf{{Ours-Dual}} (DexRep for both).

\begin{figure}[h!]
    \centering
    \vspace{-0.3cm}
    \includegraphics[ width=0.95\linewidth]{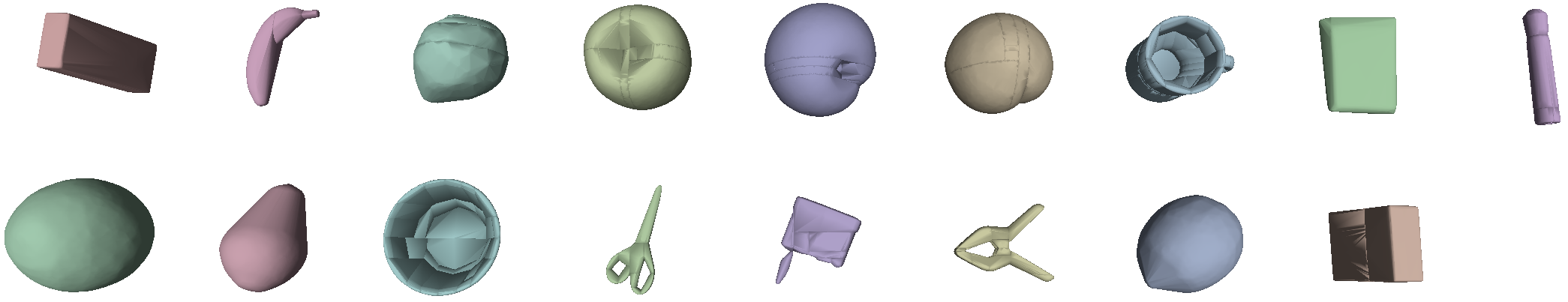}
    \caption{17 Objects used in Handover. The objects in the first row are used in training (seen), while the rest are for evaluation (unseen).}
    \label{fig:handover-test_objs}
\end{figure}

\textit{\textbf{Dataset.}}  
To evaluate generalization, we extend the single-object setup in~\cite{chen2022issacManipulation} by training with 9 YCB objects and testing on 8 unseen ones, as shown in Fig.~\ref{fig:handover-test_objs}.

\textit{\textbf{Network Architecture.}}  
The \textbf{Base} representation, also known as the robotic state $f_{prop}$, consists of the pose, velocity, joint angles, joint forces of both robotic hands, and the pose, velocity, angular velocity and target pose of the object. The single-layer FC layer processes the robotic static and DexRep inputs, with input sizes of 338, 1080, 1280, 1080, and 1280 corresponding to $f_{prop}$, $\{f_o, f_s\}$, $f_l$, $\{f_o', f_s'\}$, and $f_l'$ respectively, and an output size of 128. The policy $\pi_\theta$ is an MLP with hidden layers $[1024, 1024, 512]$ and uses the Elu activation function after each layer. The policy outputs \( \mathbf{a} \in \mathbb{R}^{20} \) containing 20 joint action.

\textit{\textbf{Training Details.}}  
DexRep is computed with voxel size $l_v = 0.02$\,m and surface range $\sigma_{max} = 0.2$\,m (Based on ablation in Sec.~\ref{Sec Ablation features parameter}). We use 4000 parallel environments and run 4000 iterations of PPO training, requiring approximately 42 GPU hours on a single RTX 3090.

\textit{\textbf{Evaluation Metric.}}  
We use the success rate and the mean reward as evaluation metrics. The handover task is considered successful when the difference between the current and target position of the object $d_{o\rightarrow t}$ is less than 0.02\,m~\cite{chen2022issacManipulation}. The mean reward is calculated as the mean cumulative reward in the most recent 200 successful manipulation trajectories.

\subsubsection{Quantitative Results and Analysis}

As shown in Table~\ref{tab:inhand&handover-test} and Fig.~\ref{fig:tasks_train} (right), \rep significantly improves the learning and performance of handover policies across all settings.

\textbf{{Ours-Dual}} achieves the highest success rate: \lqtt{{81.5\%}} on seen and \lqtt{{77.3\%}} on unseen objects, surpassing the \textbf{Base} method by \lqtt{\textbf{38.4\%}} and \lqtt{\textbf{40.5\%}}. This demonstrates the benefit of incorporating structured hand-object interaction features on both sides of a bimanual task. 

We also observe that \textbf{{Ours-Catch}} consistently outperforms \textbf{{Ours-Throw}}. This aligns with task dynamics: catching requires more precise control and tactile feedback, making high-quality contact modeling crucial. By contrast, the throwing hand mainly sets the trajectory, which is more tolerant of errors.
 
{These results affirm that even in dual-arm settings, \rep offers transferable and composable representations that enhance control performance. The consistent improvement from Base $\rightarrow$ Throw/Catch $\rightarrow$ Dual validates that the representation quality directly correlates with policy success. Moreover, the strong performance on unseen objects underscores \rep's generalization capability in tasks requiring coordinated dynamic interaction between multiple effectors.}

\section{Analyzing the Effectiveness of DexRep} \label{Sec DexRep Exp}

To comprehensively evaluate the proposed representation \rep and understand its behavior across different conditions, we design a series of experiments aimed at answering the following questions:

\begin{itemize}[label=\textbullet]
    \item \textbf{Component Contribution:} How does each component of \rep contribute to the learning performance and generalization capability in manipulation tasks?
    \item \textbf{Partial Observation Robustness:} Can DexRep maintain its effectiveness when only partial point cloud observations—common in real-world setups—are available?
    \item \textbf{Pretraining Data Efficiency:} How does the scale of pretraining data affect the feature quality and downstream performance of DexRep?
    \item \textbf{Multi-Hand Applicability:} Is DexRep adaptable to dexterous hands with different morphologies while preserving its performance?
    \item \textbf{Training Diversity Impact:} How does the number of training objects affect the generalizability of the learned policies?
    \item \textbf{Friction Robustness:} Does DexRep generalize well under varying object friction coefficients?
    \item \textbf{Feature Parameter Sensitivity:} How do different combinations of voxel edge length and maximum perception distance affect the effectiveness of DexRep across tasks?
\end{itemize}

\begin{figure}[!h]
     \centering
\vspace{-0.3cm}
    \includegraphics[trim=0 10 0 0, clip, width=0.95\linewidth]{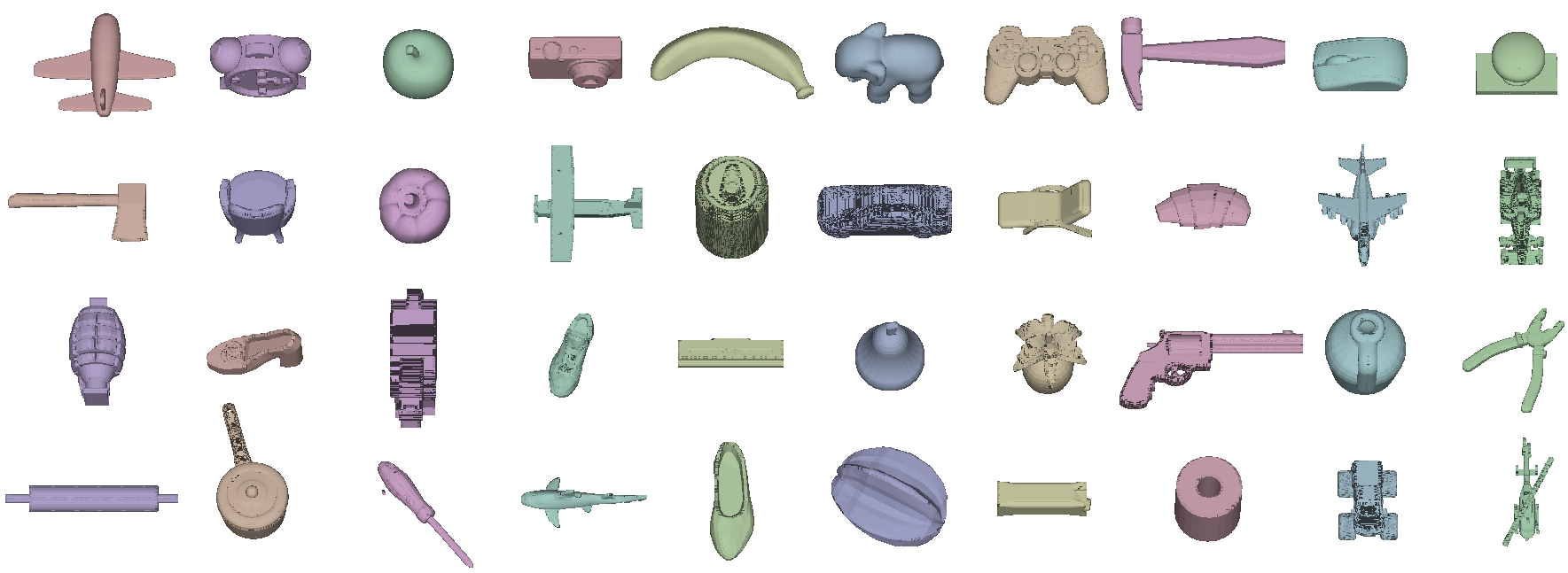}
    \caption{40 Objects used in Grasping for analyzing the effectiveness of \rep. The first 10 objects are from GRAB~\cite{taheri2020grab}, while the rest are from 3DNet~\cite{wohlkinger20123dnet}.}
   % \vskip -0.3cm
    \label{fig:grasp-test_objs}
\end{figure}

All experiments from Sec.~\ref{Sec Ablation DexRep} to Sec. \ref{Sec Multi-mor Hands} are conducted using the grasping task, with the experimental setup detailed in Section~\ref{Sec Grasp Exp Settings}. Specifically, we train policies on 40 seen objects from GRAB~\cite{taheri2020grab} and evaluate them on 10 unseen objects from GRAB and 30 unseen objects from 3DNet~\cite{wohlkinger20123dnet}, as illustrated in Fig.~\ref{fig:grasp-test_objs}. {In the following experiments, we denote the three testing object sets as \textbf{GRAB (seen), GRAB (unseen), and 3DNet (unseen)}.} The other experiments are performed under the setup in Isaac Gym. The following subsections present the results and analyses corresponding to the above questions. %{When reporting success rates in tables, ``±'' denotes the standard deviation (SD).}

\begin{figure*}
    \centering
    \includegraphics[width=0.8\linewidth]{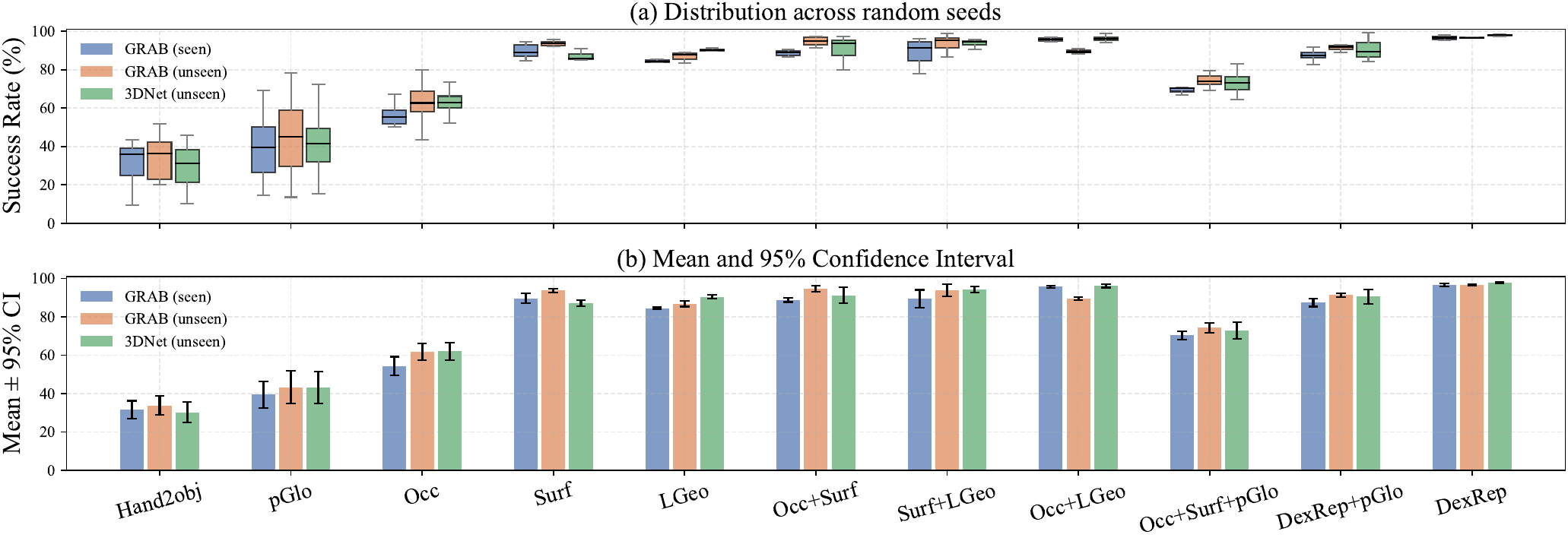}
    \vspace{-0.2cm}
    \caption{\textbf{Statistical analysis of success rates across random seeds and datasets.}
(a) Box plots illustrate the distribution of success rates over multiple training seeds for each method and dataset, where the box boundaries correspond to the 25th and 75th percentiles and the central line denotes the median (50th percentile).
(b) Bar plots report the mean success rate and 95\% confidence intervals (CIs).}
    \label{fig:ablation_dexrep_eval}
    \vspace{-0.5cm}
\end{figure*}

\begin{figure}[h!]
    \centering
    \vskip -0.2cm
         \includegraphics[width=0.8\linewidth]{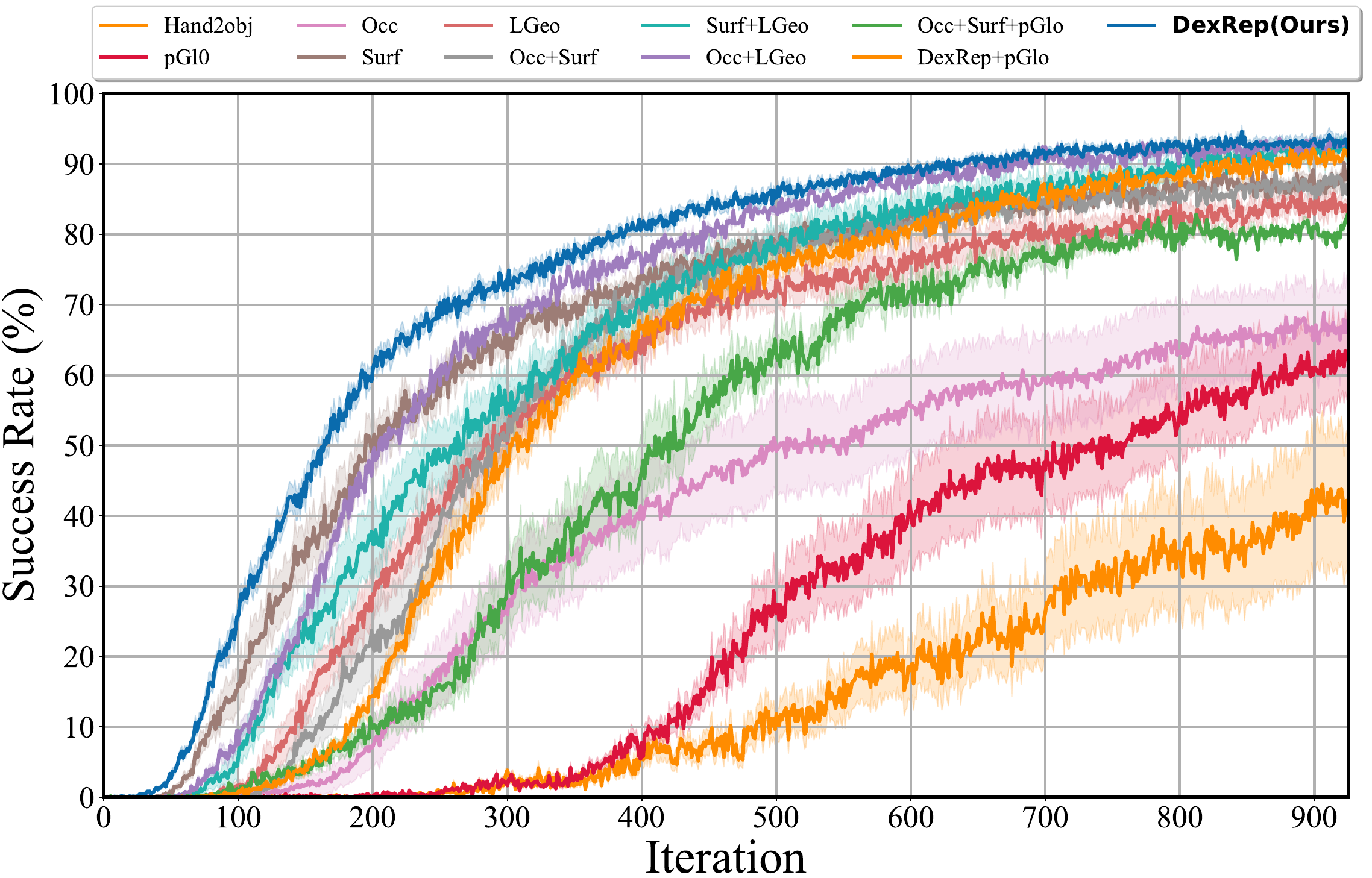}
         \vskip -0.2cm
         \caption{\lqtt{{\textbf{Success rate (\%) of our method and its variants during training.} 
         The x-axis represents the training iterations. 
         The shaded area represents the standard deviation.}} }
         \label{fig:ablation_dexrep_train}
         \vspace{-0.5cm}
\end{figure}

\subsection{Effectiveness of each component of DexRep}
\label{Sec Ablation DexRep}

We conduct ablation studies to evaluate the effectiveness of each component in DexRep and to investigate how different combinations of features influence policy learning and generalization.

\begin{itemize}[label=\textbullet]
    \item \textbf{Hand2obj:} Encodes only the relative hand-object position, following DAPG~\cite{DAPG:rajeswaran2017learning}.
    \item \textbf{pGlo:} Global shape features extracted by a pretrained PointNet~\cite{qi2017pointnet}, as used in ILAD~\cite{ILAD:wu2022learning}.
    \item \textbf{Surf}, \textbf{Occ}, \textbf{LGeo}: Variants that use only one of the proposed surface, occupancy, or local geometry features, respectively.
    \item \textbf{Surf+LGeo}, \textbf{Occ+LGeo}, \textbf{Surf+Occ}: Pairwise combinations of the above.
    \item \textbf{Occ+Surf+pGlo}: Combines the proposed features with global shape features.
    \item \textbf{DexRep+pGlo}: Adds global PointNet features to the full DexRep.
    \item \textbf{DexRep (ours)}: Full representation comprising \textbf{Surf + Occ + LGeo}.
\end{itemize}

\lqtt{The evaluation results on seen and unseen objects, together with the statistical analysis across random seeds, are presented in Fig.~\ref{fig:ablation_dexrep_eval}. 
The figure reports both the empirical distributions of success rates and the corresponding mean with 95\% confidence intervals (CIs). Non-overlapping confidence intervals indicate statistically significant performance differences ($p < 0.05$). For methods exhibiting higher variance (e.g., Hand2obj, pGlo, and Occ), we conduct 20 independent runs to obtain a more reliable estimate of their performance distribution.
Fig.~\ref{fig:ablation_dexrep_train} shows the training curves of all baselines.} 

\subsubsection{Individual Feature Effectiveness}  
{Among the single-feature variants, \textbf{Surf} achieves the highest grasping success rate on unseen objects (\lqtt{93.5\%}) and 3DNet (\lqtt{87.1\%}). This feature encodes surface distances and normals of the closest points between hand key points and the object, offering a contact-centric geometric signal that is directly relevant to manipulation. In contrast, \textbf{Occ} (coarse hand-centered occupancy) and \textbf{LGeo} (local shape patches) underperform in isolation but prove highly effective in combination with other cues. This suggests \textbf{Surf} is the most dominant standalone cue, while the others provide complementary information.}

\begin{figure*}[!ht]
     \centering
     \includegraphics[width=0.8\linewidth]{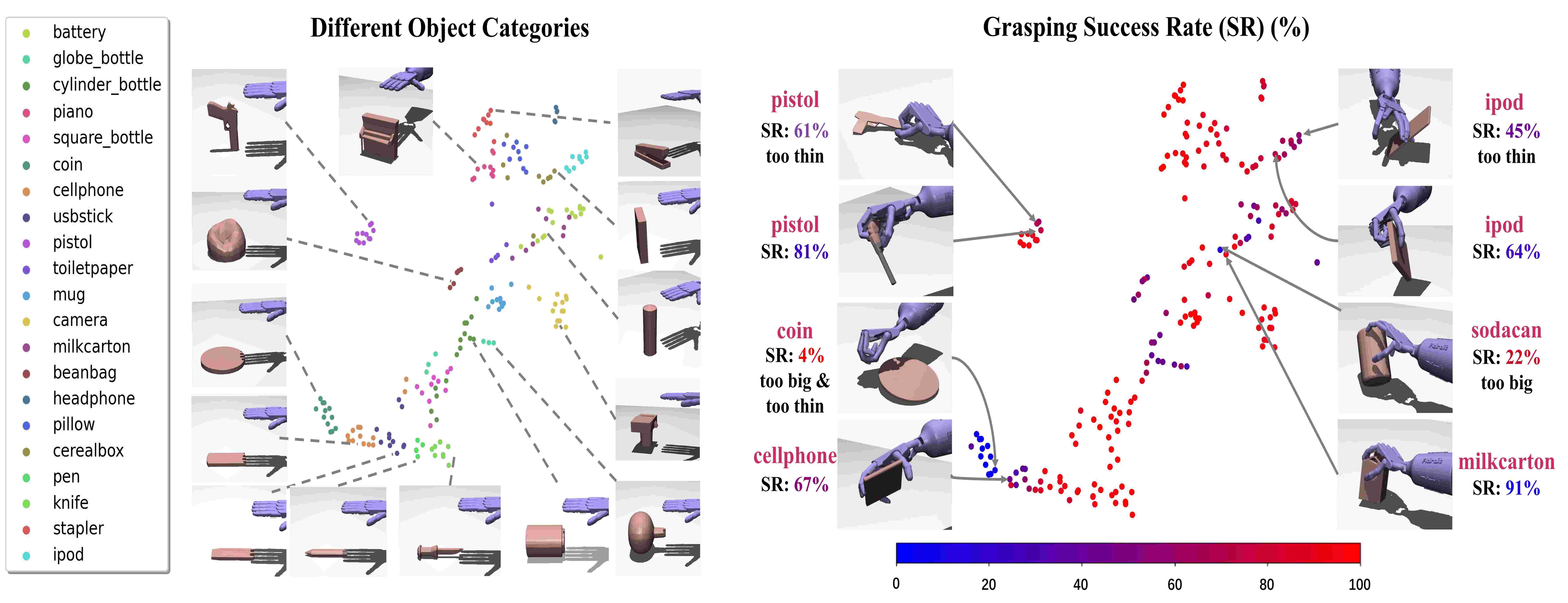}
     \caption{\textbf{T-SNE projection and grasping success rates of object categories.} Global features extracted using PointNet~\cite{qi2017pointnet} are visualized via t-SNE over objects sampled from DexGraspNet~\cite{wang2023DexGraspNet}. Color indicates object category, while grasp success rate is visualized by shading. Notably, objects with similar global features (e.g., pistol, ipod) exhibit different grasp success rates.}
     \label{fig:tsne_big}
     \vskip -0.5cm
\end{figure*}

\subsubsection{Complementarity} 
{Pairwise combinations improve performance over single-feature variants, but none match the full \textbf{DexRep} (\textbf{Surf+Occ+LGeo}), which achieves \lqtt{96.6\%} success on unseen objects and \lqtt{97.6\%} on 3DNet. The performance gap reveals how each component captures distinct and non-redundant aspects of hand-object interaction: \textbf{Surf} offers fine-grained distance-local geometry; \textbf{Occ} captures coarse volumetric occupancy relative to the hand; \textbf{LGeo} encodes more abundant local geometry via local patch descriptors. For example, \textbf{Surf+LGeo} lacks volumetric contact context (\textbf{Occ}), which may degrade performance in tasks where global pose is critical.  \textbf{Surf+Occ} has global representation and a simple surface normal descriptor, but lacks higher-order local descriptors for more complex surfaces, while the higher-order local descriptor in \textbf{LGeo+Occ} requires pretraining, which may not generalize to unseen local geometry well. Only when all three are combined does the policy consistently generalize across object categories and shape variations. \textbf{DexRep} thus forms a hierarchical spatial representation, spanning closest contact points to local and global object configurations.}

\subsubsection{Local vs. Global Representations}
When comparing \textbf{LGeo} with \textbf{pGlo}, the local representation demonstrates faster convergence during training (Fig.~\ref{fig:ablation_dexrep_train}).
From the testing results (Fig.~\ref{fig:ablation_dexrep_eval}), models using local representations also exhibit more stable performance and consistently higher success rates, exceeding \textbf{pGlo} by more than 40\%.
{Additionally, adding global object features (pGlo) consistently degrades performance. For instance, \textbf{Occ+Surf+pGlo} yields a 20\% drop on unseen objects compared to \textbf{Occ+Surf}, and \textbf{DexRep+pGlo} underperforms relative to \textbf{DexRep}. This reflects a critical insight: while global object embeddings may help in object recognition, they are often detrimental for fine-grained contact reasoning due to their sensitivity to global pose, scale, and articulation variability. In contrast, {LGeo} features—though also geometry-driven—are computed locally around anticipated contact regions, offering richer interaction-relevant cues. For example, when grasping diverse mugs or tools with similar handles but different bodies, LGeo enables the policy to generalize contact behavior across instances despite dissimilar global shapes. This validates the hypothesis that local geometry is more transferable and robust for dexterous manipulation than object-level embeddings.}

\subsubsection{Sample Efficiency}
In reinforcement learning, sample efficiency, or convergence speed, is important in dexterous manipulation tasks due to their high-dimensional action spaces and complex contact dynamics~\cite{taochen2023visualdexterity, DAPG:rajeswaran2017learning, chen2022issacManipulation,ILAD:wu2022learning}. Sample efficiency can be reflected in the number of environment interactions (i.e., samples) required for a policy to reach optimal performance. In Fig.~ \ref{fig:ablation_dexrep_train}, DexRep has converged in 600 iterations, demonstrating efficient learning compared with other representations, which remain far from convergence even at iteration 600. Particularly for the global PointNet feature, the success rate of the policy at iteration 200 is close to 0 while DexRep reaches 60\%.  While it is possible that these alternative representations may eventually converge to better performance if trained for substantially longer, the amount of computation resources required is significant. Our results highlight that DexRep achieves strong task performance with notably better sample efficiency, which is critical for the practical deployment of dexterous policies.

\subsubsection{Failure Cases}

To better understand the factors that affect robotic grasping success, we perform a t-SNE 2D projection of the global features of different object categories and analyze the failure and success cases. 
The analysis reveals that objects with similar features may exhibit different grasping success rates, as shown in Fig. \ref{fig:tsne_big}. The grasping failure cases shown in the figure (e.g., sodacan, coin, ipod, pistol) and the grasping success cases (e.g., milkcarton, cellphone, ipod, pistol) are positioned closely in the t-SNE 2D projection, yet their grasping success rates differ significantly. This indicates that even when the global features of objects are similar, specific shape details and object properties (such as size and thickness) can significantly impact the grasping success rate.

\subsection{DexRep Robustness to Partial Point Clouds} 
\label{Sec PartPoint}

Although DexRep demonstrates strong performance in both training and generalization, the above experiments assume access to complete object point clouds, which typically require multi-camera setups and careful occlusion handling in real-world scenarios. In contrast, most practical systems rely on single-view depth sensors, resulting in partial and noisy observations, especially in hand-object interaction tasks where occlusion is prominent. 

To evaluate the applicability of DexRep in such settings, we conduct experiments using partial point clouds as input. Specifically, we simulate a depth camera in MuJoCo and generate noisy partial point clouds from fixed first- or third-person viewpoints. After preprocessing, we estimate normals using the KDTreeSearchParamHybrid algorithm\footnote{\url{https://www.open3d.org/docs/latest/python_api/open3d.geometry.KDTreeSearchParamHybrid.html}}, and compute DexRep features following the same procedures as with full point clouds. We refer to these experiments with the prefix \textbf{Ours}, while the baseline uses global features extracted by a pretrained PointNet~\cite{qi2017pointnet}, labeled as \textbf{pGlo}. Specifically, \textbf{{Ours-Part-1st}} and \textbf{{pGlo-Part-1st}} use a first-person camera perspective; \textbf{{Ours-Part-3rd}} and \textbf{{pGlo-Part-3rd}} use a third-person view; \textbf{{Ours-Full}} and \textbf{{pGlo-Full}} use complete point clouds.
\begin{figure*}[!t]
    \centering
    \includegraphics[width=0.85\linewidth]{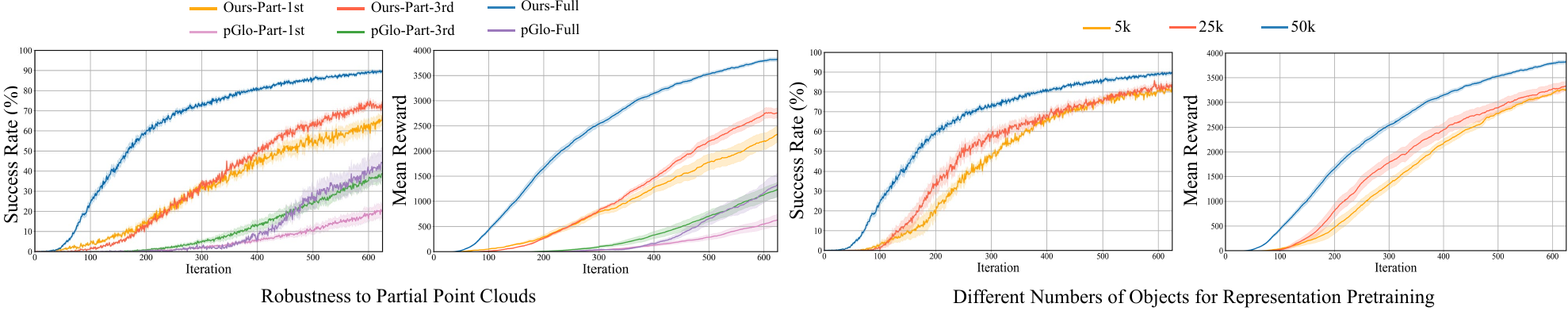}

    \vskip -0.1cm
    \caption{\lqtt{\textbf{Success rate (\%) and mean reward of our method and its variants during training.} 
    % The x-axis represents the training iterations. 
    The y-axis indicates the success rate and mean reward during training, and the shaded area represents the standard deviation.}}
    \label{fig:DexRep_Train_Curve}
    \vskip -0.5cm
\end{figure*}

\begin{figure}
    \centering
    \includegraphics[width=0.8\linewidth]{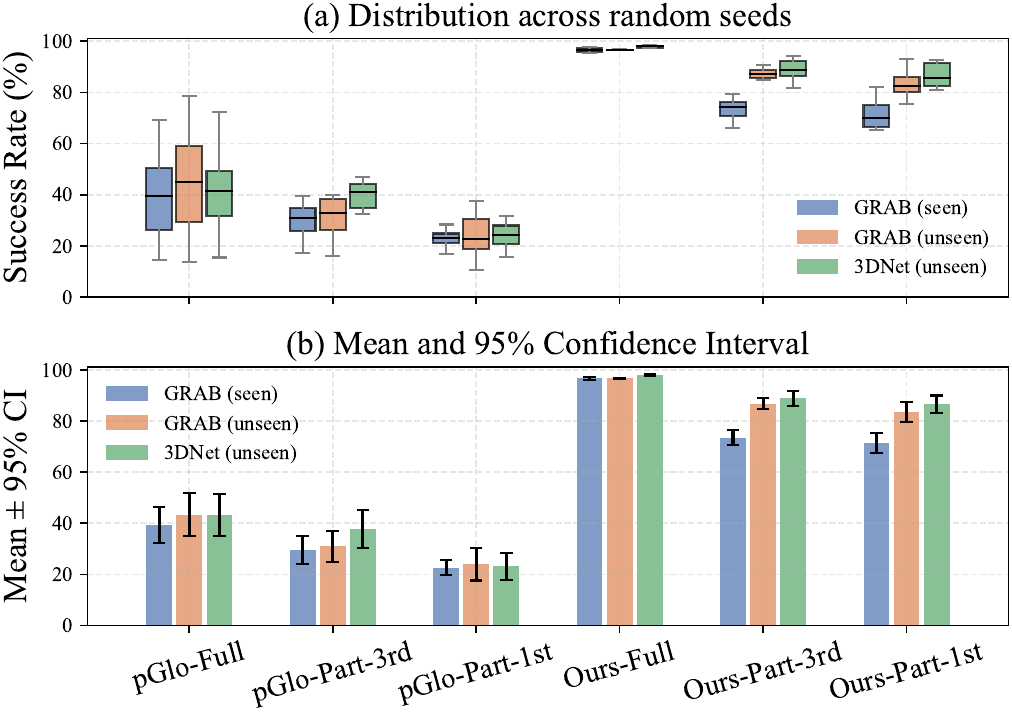}
    \caption{\textbf{Test success rate (\%) of our method applied to partial point clouds.}
We visualize both (a) the empirical distributions of success rates across random seeds and (b) the corresponding mean success rate with 95\% confidence intervals.}
    \label{fig:PartPoint}
    \vspace{-0.7cm}
\end{figure}

All experiments are trained from scratch with \rep using the pre-trained PointNet encoder. Notice that we can use a teacher-student strategy~\cite{dagger} to distill the policies using full point cloud to those using partial point cloud to improve the success rate for partial point clouds, but to demonstrate the robustness of \rep without other intervening factors, we adopt the training from scratch for a fair comparison with full point clouds. The success rate and the mean reward during the training process are shown in Fig. \ref{fig:DexRep_Train_Curve} (left). The experimental results demonstrate that \textbf{\rep} possesses a certain degree of robustness to both the completeness of point clouds and the perspective from which they are captured. The grasp success rate in the partial point cloud scenario with a fixed viewpoint can reach about 70\%. Changing the viewpoint only causes a tiny fluctuation in training. \textbf{pGlo} also shows some robustness to point cloud completeness, but is sensitive to viewpoints. 

% The final success rates are shown in Table \ref{tab:PartPoint}. 

\lqtt{We also report evaluation results with both the empirical distributions of success rates (Fig.~\ref{fig:PartPoint}(a)) and the corresponding mean success rate with 95\% confidence intervals (Fig.~\ref{fig:PartPoint}(b)) on test sets. In Fig.~\ref{fig:PartPoint}(a), the variance of our policies using different seeds is very small, indicating that all of the seeds of our policies achieve a consistently high success rate. 
% In contrast, the baselines using \textbf{pGlo} exhibit much larger variations compared with our methods, indicating instability during training. 
In Fig.~\ref{fig:PartPoint}(b), although the performance slightly decreases compared with \textbf{Ours-Full}, \textbf{Ours-Part-1st} and \textbf{Ours-Part-3rd} views achieve mean success rates of around 85\% on unseen objects. This demonstrates that our method generalizes well to partial point clouds, which can be applied to real-world robotic platforms and indicates a smaller sim-to-real gap. Additionally, it shows that the viewpoint of the point cloud has little impact on the grasp success rate, allowing flexible camera placement.}

\lqtt{For \textbf{pGlo}, the average success rates of different runs on unseen objects decrease when changing from full to partial point cloud observations. In the Fig.~\ref{fig:PartPoint}(b), the mean success rate of \textbf{pGlo-Full} is consistently higher than that of \textbf{pGlo-Part-3rd}. On the other hand, in Fig.~\ref{fig:PartPoint}(a), \textbf{{pGlo-Full}} shows high variance of different runs that some runs are higher than 60\% while some lower than 20\%, indicating less stable training. However, \textbf{pGlo-Part-3rd} maintains smaller variance.
Therefore, in addition to the comparison based on the mean success rate, we conduct independent two-sample t-tests for the results across the three datasets, which yield p-values of 0.028, 0.035, and 0.221, respectively. The p-values indicate that the difference is statistically significant for the GRAB (seen) and GRAB (unseen) datasets (p $<$ 0.05), while for 3DNet (unseen) the two methods perform comparably. We can also see from Fig.~\ref{fig:PartPoint}, the success rates in the third view (\textbf{pGlo-Part-3rd}) are higher than those in the first view, which may be attributed to the fact that the third view can observe more areas to be contacted and therefore help more in the contact reasoning when hands approach objects.}

\vspace{-0.3cm}

\subsection{Pretrain Data Volume Impact on DexRep Efficacy} \label{Sec DataV}

To evaluate how the scale of pretraining data influences the representational quality and downstream policy performance of \rep, we conduct experiments by pretraining the Loc-Geo feature extractor (PointNet~\cite{qi2017pointnet}) on \textbf{5k}, \textbf{25k}, and \textbf{50k} objects randomly sampled from ShapeNet55~\cite{chang2015shapenet}. After pretraining, we use the resulting encoders to train grasping policies on GRAB objects, and evaluate them on both seen and unseen object sets. As shown in Fig.~\ref{fig:DexRep_Train_Curve} (right), policies pretrained with larger object sets converge faster and achieve higher final rewards. Table~\ref{tab:DataVolum} summarizes the test success rates across multiple datasets.

\subsubsection{Performance Gains with More Pretraining Data}
{DexRep exhibits strong generalization even when pretrained on a relatively small dataset of 5K objects—achieving \lqtt{92.1\%} on GRAB unseen, \lqtt{92.4\%} on 3DNet, and \lqtt{77.9\%} on DexGraspNet. These results indicate that DexRep is capable of capturing transferable local geometric patterns with limited pretraining data. Across all settings, increasing the number of pretraining objects improves success rates. For instance, on GRAB seen objects, the success rate increases from \lqtt{89.3\%} with {5k} to \lqtt{96.5\%} with {50k} objects. Similar trends hold for GRAB unseen (\lqtt{92.1\%} → \lqtt{96.6\%}) and 3DNet (\lqtt{92.4\%} → \lqtt{97.6\%}). Notably, performance on the large-scale DexGraspNet benchmark improves from \lqtt{77.9\%} to \lqtt{88.1\%}. This consistent improvement confirms that larger pretraining sets enhance the geometric encoding capacity of the Loc-Geo module. A more diverse pretraining dataset allows the model to extract richer and more transferable local features, which are crucial for reliable grasp synthesis across a variety of shapes.}

\subsubsection{Stability Across Random Seeds}
{Another observation is the notable reduction in standard deviation with larger pretraining sets. For example, on GRAB seen objects, the standard deviation drops from \lqtt{4.5\%} (5k) to \lqtt{0.9\%} (50k). This indicates that larger pretraining datasets lead to more stable policies across seeds, which is essential for consistent real-world deployment.}

\begin{table}[!h]
    \centering
    % \small
    \vspace{-0.1cm}
    \caption{\textbf{Test success rate (\%) of our method with different numbers of pretraining objects.} We test on 50 objects from GRAB~\cite{taheri2020grab}, 30 unseen objects from 3DNet~\cite{wohlkinger20123dnet} and 5355 objects from DexGraspNet~\cite{wang2023DexGraspNet}.}
    % \resizebox{0.93\linewidth}{!}{
            % \setlength{\tabcolsep}{3pt}
        % \begin{threeparttable}
        {\small
        \begin{tabular}{l |c c c}
            % \toprule
            \toprule[0.5mm]
            {Testing Object Sets} & 5k & 25k & \textbf{50k} \\
            \midrule
        {GRAB (seen)}        & \lqtt{89.3{\scriptsize $\pm$4.5}} & \lqtt{90.0{\scriptsize $\pm$4.8}} & \lqtt{\textbf{96.5{\scriptsize $\pm$0.9}}} \\
        {GRAB (unseen)}      & \lqtt{92.1{\scriptsize $\pm$5.0}} & \lqtt{93.9{\scriptsize $\pm$0.9}} & \lqtt{\textbf{96.6{\scriptsize $\pm$0.4}}} \\
        {3DNet (unseen)}     & \lqtt{92.4{\scriptsize $\pm$4.0}} & \lqtt{93.7{\scriptsize $\pm$4.2}} & \lqtt{\textbf{97.6{\scriptsize $\pm$0.9}}} \\
        {DexGraspNet (unseen)} & \lqtt{77.9{\scriptsize $\pm$4.2}} & \lqtt{81.2{\scriptsize $\pm$2.9}} & \lqtt{\textbf{88.1{\scriptsize $\pm$2.0}}} \\

            \bottomrule[0.5mm]
            % \bottomrule
        \end{tabular}
        }
        % \vspace{-0.3cm}
        % \end{threeparttable}
    % }
    \label{tab:DataVolum}
\end{table}

\subsection{Influence of Multi-morphology Robotic Hands} 
\label{Sec Multi-mor Hands}

\begin{table}[htbp]
    \centering
    \tabcolsep=11.8pt
    \small
    \vspace{-0.1cm}
    \caption{\textbf{Grasp success rate (\%) on unseen objects using robotic hands with different numbers of fingers.}}
    \begin{tabular}{c|c|c|c}
        % \toprul/e
        \toprule[0.5mm]
        \# Fingers & Hand2obj & pGlo & \textbf{Ours} \\
        \midrule

        2     & \lqtt{10.2} & \lqtt{37.5} & \lqtt{\textbf{65.4}} \\
        3     & \lqtt{18.7} & \lqtt{51.1} & \lqtt{\textbf{78.2}} \\
        4     & \lqtt{15.3} & \lqtt{23.8} & \lqtt{\textbf{81.5}} \\

        \bottomrule[0.5mm]
        % \bottomrule
    \end{tabular}
    \label{tab:mulhand}
    \vskip -0.5cm
\end{table}

\begin{figure}[htbp]
    \centering
    \includegraphics[width=0.8\linewidth]{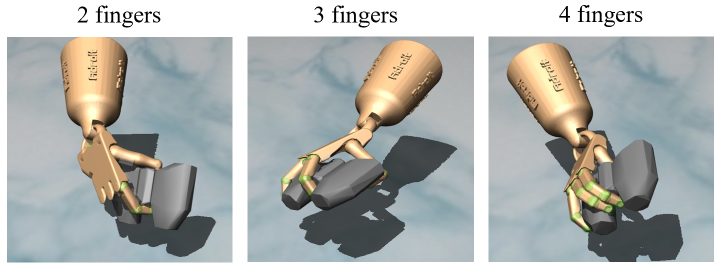}
    \caption{Examples of grasping unseen objects using robotic hands with different numbers of fingers.}
    \label{fig:diffrent morphogries}
    \vskip -0.3cm
\end{figure}

To evaluate the adaptability of \rep across robotic hands of different morphologies, we conduct experiments with a set of hands derived by selectively disassembling fingers from a standard five-fingered hand, following the procedure in~\cite{Dissemble：radosavovic2021state-only}. Examples of these hand configurations are shown in Fig.~\ref{fig:diffrent morphogries}. Table~\ref{tab:mulhand} reports the success rates of grasping unseen objects using policies trained on each hand morphology. We compare our full DexRep representation with two baselines: \textbf{Hand2obj}, which uses only relative hand-object position, and \textbf{pGlo}, which leverages global object features extracted by PointNet.

{Across all hand morphologies, our method consistently outperforms the baselines by large margins. In particular, DexRep achieves success rates of \lqtt{65.4\%} with a 2-finger hand, \lqtt{78.2\%} with a 3-finger hand, and \lqtt{81.5\%} with a 4-finger hand. These results demonstrate that DexRep generalizes effectively to different hand designs and control spaces, even when the number of available contacts is reduced. In contrast, \textbf{Hand2obj} and \textbf{pGlo} perform poorly under morphology changes, suggesting they lack the fine-grained interaction representations needed to support robust policy transfer. The strong cross-morphology performance of DexRep highlights the modularity and reusability of its representation. Since DexRep encodes interaction-relevant local geometry and spatial structure rather than hand-specific motor signals, it enables efficient policy learning even for novel hand designs.}

\vspace{-0.3cm} 

\subsection{Impact of the Number of Trained Objects in RL}
\label{Impact_of_num_trained_objs}

\begin{figure}[h!]
    \centering
    \includegraphics[width=0.8\linewidth]{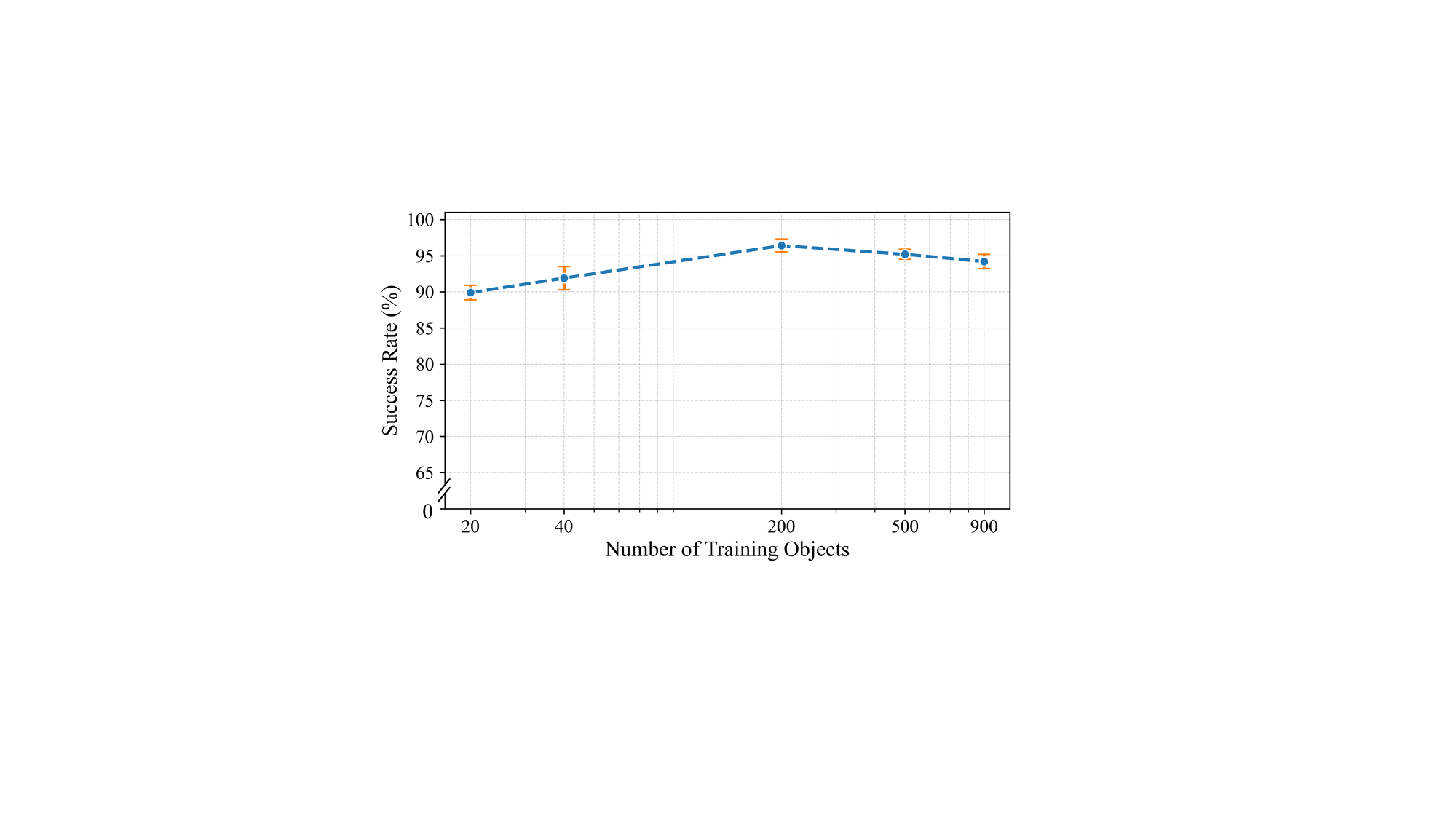}

    \caption{\lqtt{\textbf{Success rates (\%) on test objects with varying numbers of training objects.} 
    The x-axis is logarithmically scaled to represent the number of training objects. 
    The y-axis shows the mean success rate, and the error bars indicate the standard deviation}.
    }
    \label{fig:diff_num_trained_objs}
    \vspace{-0.3cm}
\end{figure}

To investigate how the diversity of training objects influences policy generalization, we migrate our learning pipeline to Isaac Gym~\cite{makoviychuk2021isaac}, which supports scalable parallel training, overcoming the limitations of MuJoCo~\cite{todorov2012mujoco} in large-scale object scenarios. We conduct reinforcement learning experiments using \rep on five object set sizes: 20, 40, 200, 500, and 900 training objects. All policies are evaluated on a fixed set of unseen test objects used in prior works~\cite{xu2023unidexgrasp,wan2023unidexgrasp++}. The results are presented in Fig.~\ref{fig:diff_num_trained_objs}.

{As the number of training objects increases, the policy's generalization performance consistently improves, indicating that object diversity enhances the robustness of learned interaction strategies. Notably, performance gains become marginal beyond 200 training objects, suggesting diminishing returns with further data scaling. This saturation implies that most of the structural variations needed for effective grasping can already be captured with a moderately sized training set. These results confirm that while large-scale object diversity is beneficial, \rep-based policies can achieve strong generalization with a relatively compact object set, making them practical for deployment in settings with limited access to extensive training assets.}

\vspace{-0.3cm} 

\subsection{Robustness to Object Friction Coefficients}
\label{Sec:FrictionRobustness}

Friction is a critical physical property that directly influences hand-object contact stability and thus the success of manipulation policies. To assess the robustness of \rep under varying contact conditions, we follow the setup of UniDexGrasp~\cite{xu2023unidexgrasp} and conduct experiments in Isaac Gym~\cite{makoviychuk2021isaac} with different object friction coefficients. We evaluate policies trained on 200 objects (based on results from Sec.~\ref{Impact_of_num_trained_objs}) using five different friction settings: fixed values of 1.0, 0.8, 0.6, 0.4, and 0.2, along with a randomized friction scenario where the coefficient is uniformly sampled in the range $[0.0, 1.0]$. The results are summarized in Table~\ref{tab:object friction coefficients}.

The results show that \rep consistently maintains high grasping success across a broad range of friction coefficients. As expected, lower friction values lead to a decline in success rates due to reduced contact stability. However, even under a challenging low-friction setting of 0.2, the policy still achieves a success rate of \lqtt{82.9\%}, reflecting strong robustness. In the random-friction setting, which simulates real-world variability, \rep attains \lqtt{90.8\%} success, only slightly lower than the performance under fixed moderate friction values. {These findings suggest that DexRep captures contact-relevant geometric cues that generalize well across diverse physical environments, reducing the need for fine-tuned physical calibration during deployment.}

\begin{table}[htbp]
  \centering
  \caption{\textbf{{Success rates (\%) on test objects under different object friction coefficients.}}}
    \begin{tabular}{cccccc}
    % \toprule
    \toprule[0.5mm]
    \multicolumn{6}{c}{Object Friction Coefficient} \\
    \midrule
    1.0 & 0.8 & 0.6 & 0.4 & 0.2 & $[0\sim1.0]$ (random) \\
    \midrule
    \lqtt{96.4} & \lqtt{93.1} & \lqtt{91.9} & \lqtt{91.5} & \lqtt{82.9} & \lqtt{90.8}\\

    % 95.9 & 93.4 & 92.2 & 91.9 & 84.7 & 91.2 \\
    % \bottomrule
    \bottomrule[0.5mm]
    \end{tabular}
    \vspace{-0.5cm}
  \label{tab:object friction coefficients}
\end{table}
\vspace{-0.3cm} 
\subsection{Ablation on Voxel Edge Length $l_v$ and Maximum Distance $\sigma_{max}$}
\label{Sec Ablation features parameter}

The proposed Surface and Occupancy features in DexRep rely on two spatial hyperparameters: the voxel edge length $l_v$ and the perception threshold $\sigma_{max}$. These parameters define the geometric resolution and spatial range of local interaction encoding. To assess their influence, we conduct an ablation study across a range of values relevant to each manipulation task. Results are shown in Table~\ref{tab:ablation_parameter}.%, with \textbf{bolded} settings indicating the default parameters used in our main experiments. 
\begin{table}[h!]
  \centering
  \scriptsize
  \tabcolsep 2.0pt
  \caption{\textbf{Success rates (\%) on unseen objects with different edge lengths of voxel $l_v$ ($m$) and the maximum distance $\sigma_{max}$ ($m$)}. The \textbf{bolded} settings are used for the main experiments.}
  \resizebox{\linewidth}{!}{
    \begin{tabular}{rrrrrrrrrr}
    \toprule[0.5mm]
    \multicolumn{10}{c}{Grasping} \\
    \midrule
    \multicolumn{1}{l}{Edge Length of Voxel $l_v$}      & 0.01    & 0.01    & 0.01    & 0.015   & 0.015   & 0.015   & 0.02    & 0.02    & \textbf{0.02} \\
    \multicolumn{1}{l}{Maximum Distance $\sigma_{max}$} & 0.1    & 0.15   & 0.2    & 0.1    & 0.15   & 0.2    & 0.1    & 0.15   & \textbf{0.2} \\

\multicolumn{1}{l}{Mean Success Rate} & \lqtt{93.4} & \lqtt{93.2} & \lqtt{92.1} & \lqtt{92.8} & \lqtt{93.7} & \lqtt{94.0} & \lqtt{94.3} & \lqtt{93.5} & \lqtt{\textbf{96.4}} \\

    \midrule[0.5mm]
    \multicolumn{10}{c}{In-Hand Orientation} \\
    \midrule
    \multicolumn{1}{l}{Edge Length of Voxel $l_v$} & \textbf{0.01}    & 0.01    & 0.01    & 0.015   & 0.015   & 0.015   & 0.02    & 0.02    & 0.02 \\
    \multicolumn{1}{l}{Maximum Distance $\sigma_{max}$} & \textbf{0.1}    & 0.15   & 0.2    & 0.1    & 0.15   & 0.2    & 0.1    & 0.15   & 0.2 \\

\multicolumn{1}{l}{Mean Success Rate} & \lqtt{\textbf{86.0}} & \lqtt{83.9} & \lqtt{76.2} & \lqtt{83.5} & \lqtt{78.1} & \lqtt{74.8} & \lqtt{84.7} & \lqtt{77.1} & \lqtt{75.3} \\

     % \multicolumn{1}{l}{Mean Success Rate}      & \textbf{85.0}   & 83.9   & 77.5  & 82.8   & 78.2   & 74.9  & 84.2   & 77.6   & 73.8  \\
    % \multicolumn{1}{l}{\color{gray} Mean success rate}      & \color{gray} \textbf{85.0}   & \color{gray} 78.6   & \color{gray} 73.6  & \color{gray} 84.3   & \color{gray} 74.1   & \color{gray} 70.4  & \color{gray} 82.8   & \color{gray} 73.6   & \color{gray} 70.7  \\
    \midrule[0.5mm]
    \multicolumn{10}{c}{Handover} \\
    \midrule
    \multicolumn{1}{l}{Edge Length of Voxel $l_v$} & 0.01    & 0.01    & 0.01    & 0.02    & \textbf{0.02}    & 0.02    & 0.03    & 0.03    & 0.03 \\
    \multicolumn{1}{l}{Maximum Distance $\sigma_{max}$} & 0.1    & 0.2    & 0.3    & 0.1    & \textbf{0.2}    & 0.3    & 0.1    & 0.2    & 0.3 \\

% \multicolumn{1}{l}{Mean Success Rate} & \lqtt{77.6} & \lqtt{76.2} & \lqtt{74.6} & \lqtt{78.3} & \lqtt{\textbf{79.1}} & \lqtt{74.1} & \lqtt{74.0} & \lqtt{73.2} & \lqtt{72.7} \\

\multicolumn{1}{l}{Mean Success Rate} & \lqtt{76.5} & \lqtt{75.6} & \lqtt{75.8} & \lqtt{75.9} & \lqtt{\textbf{77.3}} & \lqtt{75.6} & \lqtt{74.6} & \lqtt{74.7} & \lqtt{72.2} \\

    % \multicolumn{1}{l}{Mean Success Rate}        & 75.8   & 75.5   & 75.2  & 75.0   & \textbf{76.0}   & 75.5  & 74.3   & 73.4   & 72.8  \\
    % \multicolumn{1}{l}{\color{gray} Mean success rate}        & \color{gray} 70.2   & \color{gray} 70.5   & \color{gray} 71.9  & \color{gray} 74.1   & \color{gray} \textbf{76.0}   & \color{gray} 74.3  & \color{gray} 72.6   & \color{gray} 70.4   & \color{gray} 68.4  \\
    \bottomrule[0.5mm]
    \end{tabular}%                                                                
    }
  \label{tab:ablation_parameter}%
\end{table}%

\vspace{-0.3cm}

In the ablation study, the value ranges of $\sigma_{max}$ and $l_v$ are deliberately set to cover near-limit cases in the task context. For example, in the in-hand reorientation task, objects are initialized 0.1 m above the hand, and the distance between hand and object rarely exceeds 0.2 m; setting a voxel length $l_v$ of 0.03 m results in an overly coarse object volume.

As shown in Table~\ref{tab:ablation_parameter}, when these hyperparameters are within reasonable ranges, DexRep remains robust. However, two cases show clear sensitivity: (1) in the in-hand reorientation task, using a very large $\sigma_{max}$ lowers success rates because object distances greater than 0.1–0.2 m are rare, making it hard for the policy to learn from such infrequent values. Thresholding large distances to $\sigma_{max}$ reduces input variation and eases learning; also unseen values at test time are avoided, which helps the generalization.  (2) In the handover task, a very large $l_v$ degrades performance because when the voxel edge length approaches the object’s size, the global occupancy feature becomes too coarse to reliably encode the object’s pose for catching.

Based on these observations, we provide a practical guideline: for general tasks, a voxel edge length between 0.01 m and 0.02 m with a maximum perception distance $\sigma_{max}$ of about 0.1 m is recommended. For in-hand tasks that require precise local manipulation, setting $l_v$ to 0.01 m and $\sigma_{max}$ to 0.1 m is suggested to balance detail and generalization.

\color{black}

\section{Application of \rep in Real-world Scenarios} \label{Sec real world}

\subsection{System Setup and Real-world Experiment Settings}

\begin{figure}[htbp]
    \centering
    \includegraphics[width=0.8\linewidth]{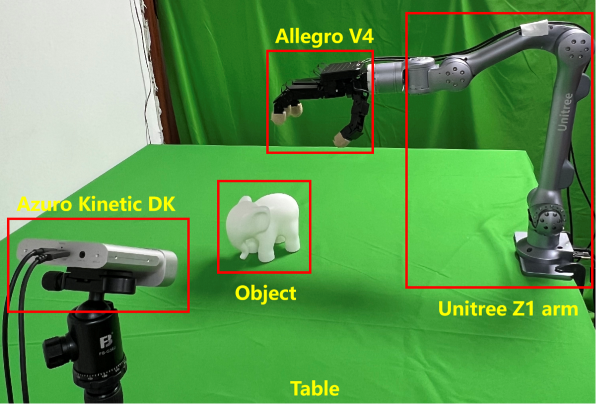}
    \caption{The grasping platform with an Allegro Hand v4, a Unitree Z1 arm and an Azure Kinect DK.}
    \vspace{-0.5cm}
    \label{fig:hardware system}
\end{figure}

To validate the effectiveness of \rep in real-world scenarios, we built a grasping system comprising an Allegro Hand\footnote{\url{https://www.wonikrobotics.com/research-robot-hand}}, a Unitree Z1 arm\footnote{\url{https://www.unitree.com/arm/}}, and an Azure Kinect DK\footnote{\url{https://azure.microsoft.com/en-us/products/kinect-dk}}, as shown in Fig.~\ref{fig:hardware system}. The joint angles of both the arm and the hand are obtained via their respective motor encoders and used to compute the spatial pose of the dexterous hand. Scene depth information is captured by the Azure Kinect DK, from which the object point cloud is segmented. This segmented point cloud, together with the estimated hand pose, is used to compute the DexRep representation. The grasping policy is trained entirely in simulation, with domain randomization applied to promote robustness. Before deployment on the real robot, we calibrate the spatial alignment between the depth camera and the robot by using a checkerboard pattern. Specifically, the checkerboard is fixed on the base of the robotic arm, enabling us to compute the transformation from the camera coordinate frame to the robot base frame. This ensures that the point cloud data from the depth camera is consistently aligned with the robot's coordinate system and the estimated hand pose. {To validate generalization to real-world objects, we evaluate the best-performing policy among three runs on eight 3D-printed objects that are not used during simulation training.} Each object is tested over 15 trials to compute the grasping success rate.

\vspace{-0.3cm}

\subsection{\rep Extraction from the Real-world Setup}

In our real-world system, \rep is computed using readily accessible sensor inputs. Specifically, the joint angles of the robotic hand, denoted by $\theta$, are obtained directly from motor encoders, while the object's point cloud $O$ (either full or partial) is captured via a depth camera. Surface normals $n_o$ of the object are estimated using the \texttt{KDTreeSearchParamHybrid} algorithm, with a search radius of 0.03 and up to 250 nearest neighbors.

Given these inputs, we employ a forward kinematics model $\mathcal{K}(\theta)$ to derive the following outputs: 1) The pose of the root joint of the middle finger, $p_{\text{mroot}}$, which serves as the spatial anchor for the Occupancy Feature; 2) The positions of predefined keypoints on the hand, denoted as $P$, which are used to compute the Surface and Local-Geo Features.

The \rep representation is then computed through the following three components:
\begin{itemize}
    \item \textbf{Occupancy Feature:}  
    A voxel grid is centered at $p_{\text{mroot}}$ to capture the local spatial occupancy of the object near the hand. Using the point cloud $O$, we determine the occupancy status of each voxel according to Eq.~\ref{OccFea_Equ}, resulting in the occupancy feature $f_o$.

    \item \textbf{Surface Feature:}  
    This component encodes the spatial relationship between the hand and the object by computing the distance and normal vectors from each hand keypoint in $P$ to its nearest object point. These are computed using $O$ and the estimated surface normals $n_o$, following Eq.~\ref{SurFea_Equ}, yielding the Surface Feature $f_s$.

    \item \textbf{Local-Geo Feature:}  
    To capture fine-grained local geometric characteristics in potential contact regions, the object point cloud $O$ is first normalized and then passed through a pretrained PointNet encoder $E(O)$. For each keypoint in $P$, the closest object point is identified, and its corresponding PointNet feature is extracted to obtain the local geometry feature $f_l$.
\end{itemize}

% \begin{figure}[htbp]
%      \centering
%      \includegraphics[width=0.48\textwidth]{fig/RealExpSuccRate.jpg}
%      \caption{Results of our real-world experiment. }
%      \label{fig:RealExp SuccRate}
%      % \vskip -0.5cm
%  \end{figure}

\vspace{-0.4cm}
\subsection{\rep with Known CAD Models}
In this experiment, we assume that object CAD models are available. Consequently, only the 6D poses of the objects are required for grasping, and these poses are obtained by registering the CAD models to the partial point clouds captured by the Azure Kinect DK's depth sensor. Specifically, objects are segmented from RGB images (recorded against a green background and calibrated with the depth sensor), and the segmented depth images are backprojected into 3D space to yield partial point clouds. To compute \rep features, we align the observed point cloud with the canonical CAD model. Specifically, we adopt the Fast Point Feature Histogram (FPFH)-based registration pipeline provided by Open3D\footnote{\url{https://www.open3d.org/html/python_api/open3d.pipelines.registration.registration_ransac_based_on_feature_matching.html\#open3d-pipelines-registration-registration-ransac-based-on-feature-matching}}.  The algorithm extracts FPFH features from both the observed partial point cloud and the model point cloud, and estimates the rigid-body transformation between them through RANSAC-based correspondence matching. The resulting transformation is used to align the CAD model to the observed object pose. To mitigate the impact of pose registration errors, we add Gaussian noise with a standard deviation of $2\,\text{cm}$ for position and $0.1\,\text{rad}$ for rotation during policy training in the MuJoCo simulator. As shown in Table \ref{tab:real-known}, our method outperforms the baselines and exhibits an outstanding generalization of unseen objects in real-world experiments under the assumption of known CAD models. 

\vspace{-0.3cm}
\subsection{\rep with Partial Point Cloud}
\label{partail}
In unstructured real-world environments, acquiring complete object point clouds is often infeasible due to occlusions and sensor limitations. To assess the robustness of \rep under partial observability, we evaluate its performance in dexterous grasping tasks using partial point clouds. Following the procedure in Sec.~\ref{Sec PartPoint}, we train grasping policies in MuJoCo using partial point clouds. To simulate real-world sensor noise, we add Gaussian noise with a standard deviation of $2~\mathrm{mm}$ to the rendered depth images during training, which are then back-projected to generate point clouds.

The trained policies are deployed on our physical system and evaluated on 8 unseen 3D-printed objects. As shown in Table~\ref{tab:real-partial}, \rep consistently outperforms baseline methods under both first-person and third-person views. However, its performance slightly degrades compared to the full point cloud scenario, highlighting the challenges posed by partial observations. To mitigate this issue, we adopt a policy distillation strategy following~\cite{taochen2023visualdexterity}, using DAgger~\cite{dagger} to transfer the full point cloud policy to one that operates on partial point clouds. This approach enables efficient skill adaptation while preserving high grasp success rates, demonstrating that \rep remains effective even under limited perceptual input.

\begin{table*}
\begin{minipage}{\linewidth}
  \centering
  \caption{\textbf{Success rate (\%) for \rep with known CAD models in the real world.}}
  \label{tab:real-known}%
    \begin{tabular}{l|ccccccccc}
    % \toprule
    \toprule[0.5mm]
    Methods & Camera & Elephant & Hand   & Binoculars & Mug    & Toothpaste & Apple  & Fryingpan & AVERAGE \\
    \midrule
    Hand2obj-CAD & 33.3   & 40.0   & 27.7   & 6.7    & 0.0    & 13.3   & 40.0   & 53.3   & 27.7  \\
    pGlo-CAD   & 86.7   & 80.0   & 60.0   & 66.7   & 33.3   & 80.0   & 60.0   & 26.7   & 61.7  \\
    Ours-CAD   & 100.0  & 66.7   & 66.7   & 86.7   & 86.7   & 93.3   & 100.0  & 60.0   & 82.5  \\
    \bottomrule[0.5mm]
    % \bottomrule
    \end{tabular}%
  
\end{minipage}%

\vspace{5.0pt}

\begin{minipage}{\linewidth}
    \centering
  \caption{\textbf{Success Rate (\%) for \rep with partial point cloud in the real world.}  Polices directly deployed from training with partial observation and distilled from full observation are included. {Ours-Full} and {pGlo-Full} (our full point cloud policy and the full point cloud baseline with global features) are provided for reference.}
    \label{tab:real-partial}
    \begin{tabular}{c|l|ccccccccc}
        % \toprule
    \toprule[0.5mm]
    \multirow{7}[6]{*}{Direct Deploy} & Methods & Camera & Elephant & Hand   & Binoculars & Mug    & Toothpaste & Apple  & Fryingpan & AVERAGE \\
\midrule          & {pGlo-Part-1st} & 13.3   & 26.7   & 20.0   & 6.7    & 26.7   & 20.0   & 20.0   & 0.0    & 16.7  \\
           & {pGlo-Part-3rd} & 26.7   & 33.3   & 26.7   & 20.0   & 33.3   & 6.7    & 40.0   & 6.7    & 24.2  \\
           \cmidrule{2-11}
           & {pGlo-Full} & 26.7   & 40.0   & 40.0   & 33.3   & 6.7    & 46.7   & 13.3   & 6.7    & 26.7  \\
\cmidrule{2-11}           & {Ours-Part-1st} & 86.7   & 93.3   & 53.3   & 80.0   & 73.3   & 60.0   & 53.3   & 26.7   & 65.8  \\
           & {Ours-Part-3rd} & 93.3   & 80.0   & 66.7   & 86.7   & 73.3   & 53.3   & 60.0   & 46.7   & 70.0  \\
           \cmidrule{2-11}
           & {Ours-Full} & 93.3   & 80.0   & 100.0  & 93.3   & 80.0   & 100.0  & 73.3   & 60.0   & 85.0  \\
    \midrule
    \multirow{2}[2]{*}{Distilled} & {Ours-Part-1st} & 100.0  & 93.3   & 86.7   & 73.3   & 66.7   & 60.0   & 73.3   & 60.0   & 76.7  \\
           & {Ours-Part-3rd} & 100.0  & 100.0  & 86.7   & 93.3   & 80.0   & 60.0   & 60.0   & 53.3   & 79.2  \\
    \bottomrule[0.5mm]
        % \bottomrule
    \end{tabular}
\end{minipage}

\vspace{5.0pt}

\begin{minipage}{\linewidth}
  \centering
  % \small
  \caption{\textbf{Success rates (\%) for comparison between \rep and Unidexgrasp++~\cite{wan2023unidexgrasp++} with an Allegro hand in Isaac Gym and real-world setup.}}
    \begin{tabular}{l|ccccccccc}
    % \topr/ule
    \toprule[0.5mm]
    Methods & Camera & Elephant & Hand   & Binoculars & Mug    & Toothpaste & Apple  & Fryingpan & AVERAGE \\
    \midrule
    {Unidexgrasp++-Full-Sim} & 93.3   & 66.7   & 73.3   & 100.0  & 66.7   & 80.0   & 60.0   & 40.0   & 72.5  \\
    {Ours-Full-Sim} & 100.0  & 93.3   & 100.0  & 93.3   & 86.7   & 100.0  & 86.7   & 60.0   & 90.0  \\
    \midrule
    {Unidexgrasp++-Full} & 86.7   & 53.3   & 66.7   & 93.3   & 40.0   & 33.3   & 40.0   & 20.0   & 54.2  \\
    {Ours-Full} & 93.3   & 80.0   & 100.0  & 93.3   & 80.0   & 100.0  & 73.3   & 60.0   & 85.0  \\
    \bottomrule[0.5mm]
    % \bottomrule
    \end{tabular}%
    \vspace{-0.5cm}
  \label{tab:real-sota}%
\end{minipage}

\end{table*}

\vspace{-0.3cm}
\subsection{Full Point Cloud Policy Deployment Compared with SOTA}

Based on the results in Table~\ref{tab:grasp-test}, UniDexGrasp++~\cite{wan2023unidexgrasp++} achieves the highest grasping success rate and is therefore selected as the baseline for our real-world evaluation. UniDexGrasp++ assumes access to a complete object point cloud; to meet this requirement, we deploy three Azure Kinect DK cameras to capture and segment the object point cloud on a tabletop setting. To ensure a fair comparison, we replicate the UniDexGrasp++ framework using the Allegro Hand in Isaac Gym and compare it against the \rep policy trained under the same conditions. During simulation training, Gaussian noise with a standard deviation of $2~\mathrm{mm}$ is added to point cloud observations, while noise with a standard deviation of 0.02 is added to the robot state—including hand joint angles, velocities, fingertip poses, global hand pose, and actions—to enhance robustness and reduce the sim-to-real gap.

We evaluate both methods on 8 unseen 3D-printed objects. As shown in Table~\ref{tab:real-sota}, our method outperforms UniDexGrasp++ by a significant margin of {30.8\% in real-world success rate}. In terms of sim-to-real transferability, UniDexGrasp++ exhibits a performance drop of approximately 18\%, while \rep only experiences a 5\% decrease. This performance gap can be largely attributed to the differing robustness to real-world noise. While UniDexGrasp++ relies heavily on precise point cloud observations, our approach utilizes a coarse voxel-based encoding for global geometry, which is inherently more tolerant to sensor noise and point cloud imperfections, leading to more stable deployment performance in the real world.

\begin{figure}[h!]
    \centering
    \vspace{-0.2cm}
    \includegraphics[width=0.85\linewidth]{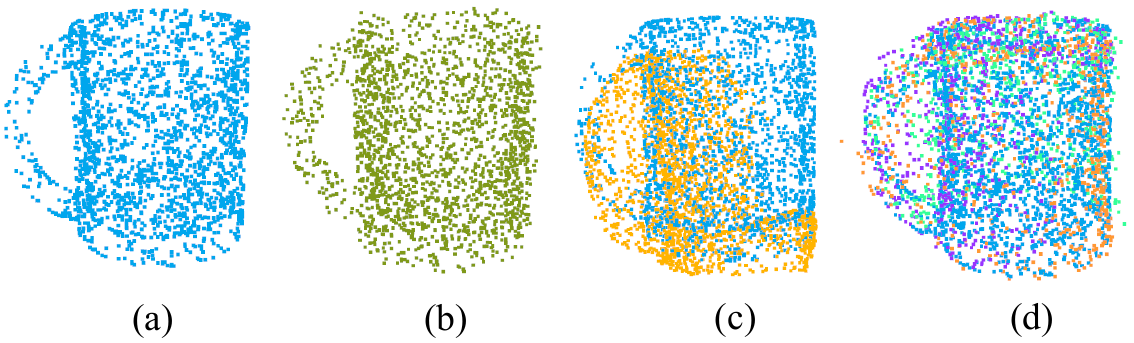}
    \caption{{\textbf{Point cloud visualization for ICP alignment and multi-camera fusion}. (a) Point clouds sampled from the CAD model of a cup (GT); (b) GT with Gaussian noise; (c) GT and partial depth point cloud after ICP registration; (d) GT and point clouds merged from multiview depth images. In (a, c, d), points in blue represent GT and point in other colors represent partial depth points from different views. }}
    \label{fig:ICP}
    \vspace{-0.5cm}
\end{figure}

\vspace{-0.2cm}

\subsection{Discussion on Different Deployment Strategies}

{An important observation arises when comparing the real-world deployment performance of different policy variants. Specifically, the success rate of the \textbf{pGlo} policy drops significantly when switching from CAD-based point clouds to full real-world point clouds (61.7\% for \textbf{pGlo-CAD} vs. 26.7\% for \textbf{{pGlo-Full}}, as shown in Table~\ref{tab:real-partial}). This substantial degradation highlights the sensitivity of global feature-based representations to real-world perception challenges, including: (1) structured, non-Gaussian noise introduced by commodity depth sensors, and (2) calibration inaccuracies inherent in multi-camera fusion setups. Fig.~\ref{fig:ICP} visualizes this discrepancy by comparing synthetic CAD-sampled point clouds (with added Gaussian noise) (b) to fused point clouds captured from a 3D-printed object using three RGB-D cameras (d). The fused real-world point clouds exhibit incomplete surfaces and inter-view inconsistencies, leading to degraded geometric fidelity. These imperfections significantly impact representations such as pGlo, which rely on clean, complete object geometry to construct meaningful global features.}

{In contrast, our DexRep-based policy demonstrates consistent performance across both point cloud modalities: both  \textbf{Ours-CAD} and \textbf{{Ours-Full}} achieve a success rate of more than  80\%. This stability reflects a core advantage of DexRep—its reliance on local, interaction-centric features rather than holistic object shape or precise CAD alignment. }

{At the same time, we notice that \textbf{{Ours-Full}} achieves a slightly higher success rate than \textbf{Ours-CAD}, 85\% vs 82.5\%, which seems contradictory to the imperfect multiview point clouds discussed. \textbf{{Ours-Full}} has calibration errors from merging multi-camera input while \textbf{Ours-CAD} has registration error. The impact of registration error for CAD models on DexRep is different from the merging error. The registration error causes a global rotation and translation misalignment from the real observation, as can be seen from Fig.~\ref{fig:ICP}(c) and the misalignment varies for different objects. When the misalignment is large, \eg~larger than 2 cm (our voxel length), the occupancy feature exhibits a large deviation from that of the ground truth point clouds and leads to a low grasp success rate for these objects. In contrast, the calibration error for the merged point clouds is the same for all objects and only a small misalignment occurs between the partial point clouds, which can be seen from the handle and edge of the cup in Fig.~\ref{fig:ICP}(d). This small misalignment can be mitigated by the occupancy feature and local feature, and therefore, \textbf{ {Ours-Full}}  maintains higher performance in real-world experiments.}

{In summary, the above results illustrate the robustness of the coarse and spatially structured encoding of DexRep: by representing geometry through voxelized occupancy and local proximity fields, DexRep maintains stable interaction representations even under significant sensor noise, occlusion, or registration error. These findings emphasize the practical resilience of DexRep in real-world manipulation scenarios. Its grounding in spatially local features enables policies to perform reliably without depending on precise CAD alignment or high-fidelity reconstructions, making DexRep particularly well-suited for deployment in unstructured or sensor-imperfect environments.}

\vspace{-0.2cm}

\subsection{Discussion: Extending to Other Tasks in the Real World}

Transferring the in-hand reorientation and handover tasks to the real world presents several additional challenges. The in-hand reorientation task and the handover task involve more complex finger coordination than the grasp task, and the latter also involves two manipulators operating in tight coordination. In these tasks, the objects are highly dynamic. Both the coordination and dynamics require high flexibility and control frequency of robotic hands for the real-world experiments, and low latency of the learned policy.  More importantly, to deploy the proposed representation, the main challenge is the severe occlusion of dynamic objects, which renders different incomplete point clouds for the same object and challenges training the policies with the partial observation from scratch. To address the challenge, a distillation solution that is used in Sec.~\ref{partail} is provided and verified in simulation. 

% Prior works such as Visual Dexterity~\cite{taochen2023visualdexterity} and Robot Synesthesia~\cite{RobotSynesthesia} have demonstrated the feasibility of in-hand manipulation using only partial visual observations by distillation. 
% Therefore, a distillation process can be used to deploy the other two tasks with DexRep in real-world settings.

% To verify it, 

{We conduct additional simulation experiments for both in-hand reorientation and handover using partial point clouds, which reflect realistic observation conditions in a real-world experiment. In the experiments, only an RGBD camera is placed in a third view. At each time step, the depth maps of objects are obtained by applying off-the-shelf segmentation techniques (e.g., SAM~\cite{kirillov2023segany}).}

%and at each time step, the depth of objects is captured and segmented using existing segmentation methods or the position of robotic hands (both can be achieved easily).

Specifically, we employ DAgger~\cite{dagger} to distill partial-observation policies using DexRep from their full-observation counterparts. For the in-hand reorientation task, the distilled policy achieves a 76.3\% success rate using partial point clouds. While the partial policy sees a drop (8.7\%) from the full policy (85.0\%), it demonstrates that meaningful control remains feasible even with sparse and noisy observations. The performance gap can be attributed to significant occlusions caused by fingers, which limit the visibility of object surfaces during manipulation. For the handover task, the occlusion is moderate, and therefore, the partial point cloud policy achieves 72.6\% success, close to the 76.0\% attained with full point cloud input.

The main reason for the high success rate after distillation to partial observation is that although occlusion is significant due to the enclosure of the finger in the in-hand orientation task, the major contribution of DexRep to the performance is the local representation. The local representation can grant the policy performance in partial observations.

\vspace{-0.2cm}

%% file: txt/conclusion.tex
\section{Conclusion}
In this paper, we propose a novel hand-object interaction representation for robotic dexterous manipulation, called \rep, which consists of the Occupancy Feature, Surface Feature, and Local-Geo Feature. 
This representation captures the relative shape feature of the objects and the spatial relation between hands and objects during hand-object interactions, which makes the learned policy generalize to novel objects well. We embed \rep into deep reinforcement learning to learn three dexterous manipulation tasks, including grasping, in-hand manipulation, and handover. We perform experiments comparing the state-of-art methods for each task. \rep achieves the highest manipulation success rates in all tasks both in simulation and real-world experiments, which demonstrates that our method can effectively capture the representation of the hand-object relationship in various dexterous manipulation tasks and generalize to unseen objects. Additionally, we conduct a thorough performance analysis of our proposed method. The experimental results demonstrate the necessity of each module in our method and show that our method can handle various dexterous manipulation scenarios.

\vspace{0.1cm}

\noindent\textbf{Limitations and Future Work.}
While our method demonstrates strong performance in simulation and real-world experiments, several limitations remain. 
First, the representation relies on object point clouds to encode hand–object interactions, which makes it less robust in cases where depth information is unreliable (e.g., transparent, reflective, or dark objects, or under severe occlusions). 
For deformable objects, the observed geometry may change drastically across time steps, reducing the stability of the learned representation.  
Second, our experiments are limited to rigid objects, leaving open the question of generalization to soft or articulated ones. 
Addressing these challenges may require integrating complementary modalities (e.g., RGB images, tactile or force sensing) to compensate for unreliable depth, as well as developing deformation-aware representations to handle shape changes. 
Finally, extending evaluations to a wider range of real-world scenarios, especially those involving deformable and visually challenging objects, remains a promising direction for future work.

%% file: root.bbl
\begin{thebibliography}{10}
\providecommand{\url}[1]{#1}
\csname url@samestyle\endcsname
\providecommand{\newblock}{\relax}
\providecommand{\bibinfo}[2]{#2}
\providecommand{\BIBentrySTDinterwordspacing}{\spaceskip=0pt\relax}
\providecommand{\BIBentryALTinterwordstretchfactor}{4}
\providecommand{\BIBentryALTinterwordspacing}{\spaceskip=\fontdimen2\font plus
\BIBentryALTinterwordstretchfactor\fontdimen3\font minus \fontdimen4\font\relax}
\providecommand{\BIBforeignlanguage}[2]{{%
\expandafter\ifx\csname l@#1\endcsname\relax
\typeout{** WARNING: IEEEtran.bst: No hyphenation pattern has been}%
\typeout{** loaded for the language `#1'. Using the pattern for}%
\typeout{** the default language instead.}%
\else
\language=\csname l@#1\endcsname
\fi
#2}}
\providecommand{\BIBdecl}{\relax}
\BIBdecl

\bibitem{DAPG:rajeswaran2017learning}
A.~Rajeswaran, V.~Kumar, A.~Gupta, G.~Vezzani, J.~Schulman, E.~Todorov, and S.~Levine, ``{Learning Complex Dexterous Manipulation with Deep Reinforcement Learning and Demonstrations},'' in \emph{Proceedings of Robotics: Science and Systems}, 2018.

\bibitem{ILAD:wu2022learning}
Y.-H. Wu, J.~Wang, and X.~Wang, ``Learning generalizable dexterous manipulation from human grasp affordance,'' in \emph{Conference on Robot Learning}.\hskip 1em plus 0.5em minus 0.4em\relax PMLR, 2023, pp. 618--629.

\bibitem{qin2022dexmv}
Y.~Qin, Y.-H. Wu, S.~Liu, H.~Jiang, R.~Yang, Y.~Fu, and X.~Wang, ``Dexmv: Imitation learning for dexterous manipulation from human videos,'' in \emph{European Conference on Computer Vision}.\hskip 1em plus 0.5em minus 0.4em\relax Springer, 2022, pp. 570--587.

\bibitem{bao2023dexart}
C.~Bao, H.~Xu, Y.~Qin, and X.~Wang, ``Dexart: Benchmarking generalizable dexterous manipulation with articulated objects,'' in \emph{Proceedings of the IEEE/CVF Conference on Computer Vision and Pattern Recognition}, 2023, pp. 21\,190--21\,200.

\bibitem{mandikal2022dexvip}
P.~Mandikal and K.~Grauman, ``Dexvip: Learning dexterous grasping with human hand pose priors from video,'' in \emph{Conference on Robot Learning}.\hskip 1em plus 0.5em minus 0.4em\relax PMLR, 2022, pp. 651--661.

\bibitem{xu2023unidexgrasp}
Y.~Xu, W.~Wan, J.~Zhang, H.~Liu, Z.~Shan, H.~Shen, R.~Wang, H.~Geng, Y.~Weng, J.~Chen \emph{et~al.}, ``Unidexgrasp: Universal robotic dexterous grasping via learning diverse proposal generation and goal-conditioned policy,'' in \emph{Proceedings of the IEEE/CVF Conference on Computer Vision and Pattern Recognition}, 2023, pp. 4737--4746.

\bibitem{wang2023DexGraspNet}
R.~Wang, J.~Zhang, J.~Chen, Y.~Xu, P.~Li, T.~Liu, and H.~Wang, ``Dexgraspnet: A large-scale robotic dexterous grasp dataset for general objects based on simulation,'' in \emph{2023 IEEE International Conference on Robotics and Automation (ICRA)}.\hskip 1em plus 0.5em minus 0.4em\relax IEEE, 2023, pp. 11\,359--11\,366.

\bibitem{wan2023unidexgrasp++}
W.~Wan, H.~Geng, Y.~Liu, Z.~Shan, Y.~Yang, L.~Yi, and H.~Wang, ``Unidexgrasp++: Improving dexterous grasping policy learning via geometry-aware curriculum and iterative generalist-specialist learning,'' in \emph{Proceedings of the IEEE/CVF International Conference on Computer Vision}, 2023, pp. 3891--3902.

\bibitem{taochen2023visualdexterity}
T.~Chen, M.~Tippur, S.~Wu, V.~Kumar, E.~Adelson, and P.~Agrawal, ``Visual dexterity: In-hand reorientation of novel and complex object shapes,'' \emph{Science Robotics}, vol.~8, no.~84, p. eadc9244, 2023.

\bibitem{bi-dexhands}
Y.~Chen, Y.~Geng, F.~Zhong, J.~Ji, J.~Jiang, Z.~Lu, H.~Dong, and Y.~Yang, ``Bi-dexhands: Towards human-level bimanual dexterous manipulation,'' \emph{IEEE Transactions on Pattern Analysis and Machine Intelligence}, vol.~46, no.~5, pp. 2804--2818, 2024.

\bibitem{li2023dexdeform}
S.~Li, Z.~Huang, T.~Chen, T.~Du, H.~Su, J.~B. Tenenbaum, and C.~Gan, ``Dexdeform: Dexterous deformable object manipulation with human demonstrations and differentiable physics,'' in \emph{The Eleventh International Conference on Learning Representations}.

\bibitem{allegrohand}
Allegro hand. \url{https://www.wonikrobotics.com/robot-hand}.

\bibitem{shaw2023leap}
K.~Shaw, A.~Agarwal, and D.~Pathak, ``{LEAP Hand: Low-Cost, Efficient, and Anthropomorphic Hand for Robot Learning},'' in \emph{Proceedings of Robotics: Science and Systems}, Daegu, Republic of Korea, July 2023.

\bibitem{shadowrobot_hand}
{Shadow Robot}. (2005) Shadowrobot dexterous hand. \url{https://www.shadowrobot.com/dexterous-hand-series/}.

\bibitem{qi2017pointnet}
R.~Q. Charles, H.~Su, M.~Kaichun, and L.~J. Guibas, ``Pointnet: Deep learning on point sets for 3d classification and segmentation,'' in \emph{2017 IEEE Conference on Computer Vision and Pattern Recognition (CVPR)}.\hskip 1em plus 0.5em minus 0.4em\relax IEEE, 2017, pp. 77--85.

\bibitem{feix2016taxonomy}
T.~Feix, J.~Romero, H.-B. Schmiedmayer, A.~M. Dollar, and D.~Kragic, ``The grasp taxonomy of human grasp types,'' \emph{IEEE Transactions on Human-Machine Systems}, vol.~46, no.~1, pp. 66--77, 2016.

\bibitem{mason2001hand}
C.~R. Mason, J.~E. Gomez, and T.~J. Ebner, ``Hand synergies during reach-to-grasp,'' \emph{Journal of neurophysiology}, vol.~86, no.~6, pp. 2896--2910, 2001.

\bibitem{liu2023dexrepnet}
Q.~Liu, Y.~Cui, Q.~Ye, Z.~Sun, H.~Li, G.~Li, L.~Shao, and J.~Chen, ``Dexrepnet: Learning dexterous robotic grasping network with geometric and spatial hand-object representations,'' in \emph{2023 IEEE/RSJ International Conference on Intelligent Robots and Systems (IROS)}.\hskip 1em plus 0.5em minus 0.4em\relax IEEE, 2023, pp. 3153--3160.

\bibitem{han1998dextrous}
L.~Han and J.~C. Trinkle, ``Dextrous manipulation by rolling and finger gaiting,'' in \emph{Proceedings. 1998 IEEE International Conference on Robotics and Automation (Cat. No. 98CH36146)}, vol.~1.\hskip 1em plus 0.5em minus 0.4em\relax IEEE, 1998, pp. 730--735.

\bibitem{rus1999hand}
D.~Rus, ``In-hand dexterous manipulation of piecewise-smooth 3-d objects,'' \emph{The International Journal of Robotics Research}, vol.~18, no.~4, pp. 355--381, 1999.

\bibitem{mordatch2012contact}
I.~Mordatch, Z.~Popovi{\'c}, and E.~Todorov, ``Contact-invariant optimization for hand manipulation,'' in \emph{Proceedings of the ACM SIGGRAPH/Eurographics symposium on computer animation}, 2012, pp. 137--144.

\bibitem{kumar2014real}
V.~Kumar, Y.~Tassa, T.~Erez, and E.~Todorov, ``Real-time behaviour synthesis for dynamic hand-manipulation,'' in \emph{2014 IEEE International Conference on Robotics and Automation (ICRA)}.\hskip 1em plus 0.5em minus 0.4em\relax IEEE, 2014, pp. 6808--6815.

\bibitem{GRAFF:mandikal2021learning}
P.~Mandikal and K.~Grauman, ``Learning dexterous grasping with object-centric visual affordances,'' in \emph{2021 IEEE international conference on robotics and automation (ICRA)}.\hskip 1em plus 0.5em minus 0.4em\relax IEEE, 2021, pp. 6169--6176.

\bibitem{Christen_2022_CVPR}
S.~Christen, M.~Kocabas, E.~Aksan, J.~Hwangbo, J.~Song, and O.~Hilliges, ``D-grasp: Physically plausible dynamic grasp synthesis for hand-object interactions,'' in \emph{Proceedings of the IEEE/CVF Conference on Computer Vision and Pattern Recognition}, 2022, pp. 20\,577--20\,586.

\bibitem{yang2022chopsticks}
Z.~Yang, K.~Yin, and L.~Liu, ``Learning to use chopsticks in diverse gripping styles,'' \emph{ACM Transactions on Graphics (TOG)}, vol.~41, no.~4, pp. 1--17, 2022.

\bibitem{lan2023dexcatch}
F.~Lan, S.~Wang, Y.~Zhang, H.~Xu, O.~O. Oseni, Z.~Zhang, Y.~Gao, and T.~Zhang, ``Dexcatch: Learning to catch arbitrary objects with dexterous hands,'' in \emph{8th Annual Conference on Robot Learning}.

\bibitem{huang2021generalization}
W.~Huang, I.~Mordatch, P.~Abbeel, and D.~Pathak, ``Generalization in dexterous manipulation via geometry-aware multi-task learning,'' \emph{arXiv preprint arXiv:2111.03062}, 2021.

\bibitem{chen2022inhand}
T.~Chen, J.~Xu, and P.~Agrawal, ``A system for general in-hand object re-orientation,'' in \emph{Conference on Robot Learning}, 2022, pp. 297--307.

\bibitem{qi2023general}
H.~Qi, B.~Yi, S.~Suresh, M.~Lambeta, Y.~Ma, R.~Calandra, and J.~Malik, ``General in-hand object rotation with vision and touch,'' in \emph{Conference on Robot Learning}, 2023, pp. 2549--2564.

\bibitem{omer2021model}
M.~Omer, R.~Ahmed, B.~Rosman, and S.~F. Babikir, ``Model predictive-actor critic reinforcement learning for dexterous manipulation,'' in \emph{2020 International Conference on Computer, Control, Electrical, and Electronics Engineering (ICCCEEE)}.\hskip 1em plus 0.5em minus 0.4em\relax IEEE, 2021, pp. 1--6.

\bibitem{he2022discovering}
Z.~He and M.~Ciocarlie, ``Discovering synergies for robot manipulation with multi-task reinforcement learning,'' in \emph{2022 International Conference on Robotics and Automation (ICRA)}.\hskip 1em plus 0.5em minus 0.4em\relax IEEE, 2022, pp. 2714--2721.

\bibitem{qin2022one}
Y.~Qin, H.~Su, and X.~Wang, ``From one hand to multiple hands: Imitation learning for dexterous manipulation from single-camera teleoperation,'' \emph{IEEE Robotics and Automation Letters}, vol.~7, no.~4, pp. 10\,873--10\,881, 2022.

\bibitem{she2022ibs}
Q.~She, R.~Hu, J.~Xu, M.~Liu, K.~Xu, and H.~Huang, ``Learning high-dof reaching-and-grasping via dynamic representation of gripper-object interaction,'' \emph{ACM Transactions on Graphics (TOG)}, vol.~41, no.~4, pp. 1--14, 2022.

\bibitem{levine2018handeye}
S.~Levine, P.~Pastor, A.~Krizhevsky, J.~Ibarz, and D.~Quillen, ``Learning hand-eye coordination for robotic grasping with deep learning and large-scale data collection,'' \emph{The International Journal of Robotics Research}, 2018.

\bibitem{liu2020deep}
M.~Liu, Z.~Pan, K.~Xu, K.~Ganguly, and D.~Manocha, ``Deep differentiable grasp planner for high-dof grippers,'' in \emph{Robotics: Science and Systems}, 2020.

\bibitem{wei2022dvgg}
W.~Wei, D.~Li, P.~Wang, Y.~Li, W.~Li, Y.~Luo, and J.~Zhong, ``Dvgg: Deep variational grasp generation for dextrous manipulation,'' \emph{IEEE Robotics and Automation Letters}, vol.~7, no.~2, pp. 1659--1666, 2022.

\bibitem{qin2023dexpoint}
Y.~Qin, B.~Huang, Z.-H. Yin, H.~Su, and X.~Wang, ``Dexpoint: Generalizable point cloud reinforcement learning for sim-to-real dexterous manipulation,'' in \emph{Conference on Robot Learning}.\hskip 1em plus 0.5em minus 0.4em\relax PMLR, 2023, pp. 594--605.

\bibitem{varley2017shape}
J.~Varley, C.~DeChant, A.~Richardson, J.~Ruales, and P.~Allen, ``Shape completion enabled robotic grasping,'' in \emph{2017 IEEE/RSJ international conference on intelligent robots and systems (IROS)}.\hskip 1em plus 0.5em minus 0.4em\relax IEEE, 2017, pp. 2442--2447.

\bibitem{cao2021suctionnet}
H.~Cao, H.-S. Fang, W.~Liu, and C.~Lu, ``Suctionnet-1billion: A large-scale benchmark for suction grasping,'' \emph{IEEE Robotics and Automation Letters}, vol.~6, no.~4, pp. 8718--8725, 2021.

\bibitem{nair2022r3m}
S.~Nair, A.~Rajeswaran, V.~Kumar, C.~Finn, and A.~Gupta, ``R3m: A universal visual representation for robot manipulation,'' in \emph{Conference on Robot Learning}.\hskip 1em plus 0.5em minus 0.4em\relax PMLR, 2023, pp. 892--909.

\bibitem{radosavovic2023real}
I.~Radosavovic, T.~Xiao, S.~James, P.~Abbeel, J.~Malik, and T.~Darrell, ``Real-world robot learning with masked visual pre-training,'' in \emph{Conference on Robot Learning}.\hskip 1em plus 0.5em minus 0.4em\relax PMLR, 2023, pp. 416--426.

\bibitem{he2022mae}
K.~He, X.~Chen, S.~Xie, Y.~Li, P.~Doll{\'a}r, and R.~Girshick, ``Masked autoencoders are scalable vision learners,'' in \emph{Proceedings of the IEEE/CVF conference on computer vision and pattern recognition}, 2022, pp. 16\,000--16\,009.

\bibitem{shao2020unigrasp}
L.~Shao, F.~Ferreira, M.~Jorda, V.~Nambiar, J.~Luo, E.~Solowjow, J.~A. Ojea, O.~Khatib, and J.~Bohg, ``Unigrasp: Learning a unified model to grasp with multifingered robotic hands,'' \emph{IEEE Robotics and Automation Letters}, vol.~5, no.~2, pp. 2286--2293, 2020.

\bibitem{joshi2020robotic}
S.~Joshi, S.~Kumra, and F.~Sahin, ``Robotic grasping using deep reinforcement learning,'' in \emph{2020 IEEE 16th International Conference on Automation Science and Engineering (CASE)}.\hskip 1em plus 0.5em minus 0.4em\relax IEEE, 2020, pp. 1461--1466.

\bibitem{chen2022dextransfer}
Z.~Q. Chen, K.~Van~Wyk, Y.-W. Chao, W.~Yang, A.~Mousavian, A.~Gupta, and D.~Fox, ``Dextransfer: Real world multi-fingered dexterous grasping with minimal human demonstrations,'' \emph{arXiv preprint arXiv:2209.14284}, 2022.

\bibitem{zhang2021manipnet}
H.~Zhang, Y.~Ye, T.~Shiratori, and T.~Komura, ``Manipnet: neural manipulation synthesis with a hand-object spatial representation,'' \emph{ACM Transactions on Graphics (ToG)}, vol.~40, no.~4, pp. 1--14, 2021.

\bibitem{chang2015shapenet}
A.~X. Chang, T.~Funkhouser, L.~Guibas, P.~Hanrahan, Q.~Huang, Z.~Li, S.~Savarese, M.~Savva, S.~Song, H.~Su \emph{et~al.}, ``Shapenet: An information-rich 3d model repository,'' \emph{arXiv preprint arXiv:1512.03012}, 2015.

\bibitem{achlioptas2018learning}
P.~Achlioptas, O.~Diamanti, I.~Mitliagkas, and L.~Guibas, ``Learning representations and generative models for 3d point clouds,'' in \emph{International conference on machine learning}.\hskip 1em plus 0.5em minus 0.4em\relax PMLR, 2018, pp. 40--49.

\bibitem{fan2017point}
H.~Fan, H.~Su, and L.~J. Guibas, ``A point set generation network for 3d object reconstruction from a single image,'' in \emph{Proceedings of the IEEE conference on computer vision and pattern recognition}, 2017, pp. 605--613.

\bibitem{taheri2020grab}
O.~Taheri, N.~Ghorbani, M.~J. Black, and D.~Tzionas, ``Grab: A dataset of whole-body human grasping of objects,'' in \emph{European conference on computer vision}.\hskip 1em plus 0.5em minus 0.4em\relax Springer, 2020, pp. 581--600.

\bibitem{kumar2013adroithand}
V.~Kumar, Z.~Xu, and E.~Todorov, ``Fast, strong and compliant pneumatic actuation for dexterous tendon-driven hands,'' in \emph{2013 IEEE international conference on robotics and automation}.\hskip 1em plus 0.5em minus 0.4em\relax IEEE, 2013, pp. 1512--1519.

\bibitem{chen2022issacManipulation}
Y.~Chen, T.~Wu, S.~Wang, X.~Feng, J.~Jiang, Z.~Lu, S.~McAleer, H.~Dong, S.-C. Zhu, and Y.~Yang, ``Towards human-level bimanual dexterous manipulation with reinforcement learning,'' \emph{Advances in Neural Information Processing Systems}, vol.~35, pp. 5150--5163, 2022.

\bibitem{MANO:SIGGRAPHASIA:2017}
J.~Romero, D.~Tzionas, and M.~J. Black, ``Embodied hands: Modeling and capturing hands and bodies together,'' \emph{ACM Transactions on Graphics, (Proc. SIGGRAPH Asia)}, vol.~36, no.~6, Nov. 2017.

\bibitem{handa2020dexpilot}
A.~Handa, K.~Van~Wyk, W.~Yang, J.~Liang, Y.-W. Chao, Q.~Wan, S.~Birchfield, N.~Ratliff, and D.~Fox, ``Dexpilot: Vision-based teleoperation of dexterous robotic hand-arm system,'' in \emph{2020 IEEE International Conference on Robotics and Automation (ICRA)}.\hskip 1em plus 0.5em minus 0.4em\relax IEEE, 2020, pp. 9164--9170.

\bibitem{todorov2012mujoco}
E.~Todorov, T.~Erez, and Y.~Tassa, ``Mujoco: A physics engine for model-based control,'' in \emph{2012 IEEE/RSJ international conference on intelligent robots and systems}.\hskip 1em plus 0.5em minus 0.4em\relax IEEE, 2012, pp. 5026--5033.

\bibitem{GAE:schulman2015high}
J.~Schulman, P.~Moritz, S.~Levine, M.~Jordan, and P.~Abbeel, ``High-dimensional continuous control using generalized advantage estimation,'' \emph{arXiv preprint arXiv:1506.02438}, 2015.

\bibitem{schulman2017ppo}
J.~Schulman, F.~Wolski, P.~Dhariwal, A.~Radford, and O.~Klimov, ``Proximal policy optimization algorithms,'' \emph{arXiv preprint arXiv:1707.06347}, 2017.

\bibitem{makoviychuk2021isaac}
V.~Makoviychuk, L.~Wawrzyniak, Y.~Guo, M.~Lu, K.~Storey, M.~Macklin, D.~Hoeller, N.~Rudin, A.~Allshire, A.~Handa \emph{et~al.}, ``Isaac gym: High performance gpu based physics simulation for robot learning,'' in \emph{Thirty-fifth Conference on Neural Information Processing Systems Datasets and Benchmarks Track (Round 2)}.

\bibitem{calli2017ycb}
B.~Calli, A.~Singh, J.~Bruce, A.~Walsman, K.~Konolige, S.~Srinivasa, P.~Abbeel, and A.~M. Dollar, ``Yale-cmu-berkeley dataset for robotic manipulation research,'' \emph{The International Journal of Robotics Research}, vol.~36, no.~3, pp. 261--268, 2017.

\bibitem{brahmbhatt2019contactdb}
S.~Brahmbhatt, C.~Ham, C.~C. Kemp, and J.~Hays, ``Contactdb: Analyzing and predicting grasp contact via thermal imaging,'' in \emph{Proceedings of the IEEE/CVF conference on computer vision and pattern recognition}, 2019, pp. 8709--8719.

\bibitem{wohlkinger20123dnet}
W.~Wohlkinger, A.~Aldoma, R.~B. Rusu, and M.~Vincze, ``3dnet: Large-scale object class recognition from cad models,'' in \emph{2012 IEEE international conference on robotics and automation}.\hskip 1em plus 0.5em minus 0.4em\relax IEEE, 2012, pp. 5384--5391.

\bibitem{dagger}
S.~Ross, G.~Gordon, and D.~Bagnell, ``A reduction of imitation learning and structured prediction to no-regret online learning,'' in \emph{Proceedings of the fourteenth international conference on artificial intelligence and statistics}.\hskip 1em plus 0.5em minus 0.4em\relax JMLR Workshop and Conference Proceedings, 2011, pp. 627--635.

\bibitem{Dissemble：radosavovic2021state-only}
I.~Radosavovic, X.~Wang, L.~Pinto, and J.~Malik, ``State-only imitation learning for dexterous manipulation,'' in \emph{2021 IEEE/RSJ International Conference on Intelligent Robots and Systems (IROS)}.\hskip 1em plus 0.5em minus 0.4em\relax IEEE, 2021, pp. 7865--7871.

\bibitem{kirillov2023segany}
A.~Kirillov, E.~Mintun, N.~Ravi, H.~Mao, C.~Rolland, L.~Gustafson, T.~Xiao, S.~Whitehead, A.~C. Berg, W.-Y. Lo, P.~Doll{\'a}r, and R.~Girshick, ``Segment anything,'' \emph{arXiv:2304.02643}, 2023.

\end{thebibliography}
